\documentclass[10pt,leqno]{amsart}
\usepackage{titlesec} 
\usepackage{graphicx}
\usepackage{adjustbox}
\usepackage{indentfirst,csquotes}

\usepackage{newtxtext}       
\usepackage{newtxmath}       
\usepackage{courier}          

\titleformat{\section}
 {\normalfont\fontfamily{ptm}\bfseries\large}
  {\thesection} 
  {1em} 
  {} 

\titleformat{\subsection}
  {\normalfont\fontfamily{ptm}\bfseries} 
  {} 
  {0em} 
  {\bfseries \thesubsection\quad} 

\titleformat{\subsubsection}
  [runin] 
  {\normalfont\fontfamily{ptm}\bfseries} 
  {\thesubsubsection} 
  {0.5em} 
  {} 
  [\quad] 

\usepackage[colorlinks = true,
            linkcolor = black,
            filecolor = black,
            urlcolor = black,
            citecolor = black,
            breaklinks = true]{hyperref}
\makeatletter

\usepackage{amsthm,amsmath} 
\usepackage{xcolor,paralist,fancyhdr,etoolbox}
\usepackage{booktabs}
\usepackage{multirow}
\usepackage{array}
\usepackage{setspace} 
\usepackage{ragged2e} 
\usepackage{tabularx}

\makeatletter
\def\@author{} 
\renewcommand{\author}[1]{%
  \g@addto@macro\@author{\def\and{, }
  #1\global\let\and\relax}
}

\renewcommand{\@maketitle}{%
    \newpage
    \null
    \vskip 2em%
    \begin{center}%
        \let \footnote \thanks
        {\LARGE \@title \par}%
        \vskip 1.5em%
        {\large
            \lineskip .5em%
            \begin{tabular}[t]{c}%
                \@author
            \end{tabular}\par}%
        \vskip 1em%
        {\large \@date}%
    \end{center}%
    \par
    \vskip 1.5em
    \@setabstract
}
\makeatother

\usepackage{biblatex} 
\DeclareFieldFormat{url}{\newline\url{#1}} 
\begin{document}

\title{TriAlignXA: An Explainable Trilemma Alignment Framework for Trustworthy Agri-product Grading}
\author{Jianfei Xie \and Ziyang Li}
\date{\today}
\address{School of Software, Xinjiang University, Urumqi 830002, China}
\email{xiejianffei@163.com}

\begin{abstract}
The 'trust deficit' in online fruit and vegetable e-commerce stems from the inability of digital transactions to provide direct sensory perception of agricultural product quality. Based on this, this paper constructs a 'Trust Pyramid' model through 'dual-source verification' of consumer trust. Experimental results confirm that quality is the cornerstone of trust. The study further reveals an 'impossible triangle' in agricultural product grading, composed of biological characteristics, timeliness, and economic viability, highlighting the limitations of traditional absolute grading standards.

To quantitatively assess this complex trade-off, this paper introduces the 'Triangular Trust Index'(TTI). Building upon this, we redefine the role of algorithms from 'decision-makers' to 'providers of transparent decision-making bases' thereby designing the explainable AI framework—TriAlignXA. This framework aims to provide decision support for a trustworthy online transaction ecosystem within the constraints of the agricultural 'impossible triangle' through multi-objective collaborative trade-offs. Its core operation relies on three engines: the Bio-Adaptive Engine handles granular quality description; the Timeliness Optimization Engine focuses on enhancing processing efficiency; and the Economic Optimization Engine controls system costs. Concurrently, the framework's integrated "Pre-Mapping Mechanism" encodes process data into QR codes, transparently conveying quality information to consumers.

To validate TriAlignXA's practical efficacy, we applied it to grading tasks. Experiments demonstrated significantly higher accuracy than baseline models. This empirical evidence, combined with theoretical analysis, jointly verifies the framework's exceptional balancing capability in addressing the "impossible triangle." In summary, this research provides comprehensive support---from theory to practice---for building a trustworthy online fruit and vegetable e-commerce ecosystem, establishing a critical pathway from algorithmic decision-making to consumer trust.

\textbf{Keywords:} Trust-Building, Explainable-AI, Impossible-Triangle, Triangular Trust Index, Pre-mapping Mechanism
\end{abstract}
\maketitle 

\bigskip

\section{Introduction}
The global expansion and deepening of the internet are fundamentally transforming traditional marketing models \cite{1}. Concurrent with this transformation, China's fruit and vegetable e-commerce market, regarded as a paradigm of the digital economy, is undergoing robust growth. Nevertheless, the prevailing distrust poses a substantial impediment to its continued growth \cite{2}. Trust is of the essence in e-commerce, impacting consumer purchasing decisions and maintaining the viability of the ecosystem \cite{3,4}. In order to address the challenges related to trust, it is imperative to undertake a thorough examination of the underlying causes and to formulate effective solutions. This study utilizes China as a model to elucidate issues of a similar nature that may arise in other regions. The extant research is principally divided into two approaches, yet neither has systematically resolved the trust crisis.

\begin{itemize}
  \item \textbf{Indirect Trust Verification Based on Additional Attributes}
  
  This approach fosters trust through the use of peripheral evidence, such as external product information, including blockchain traceability for origin verification, reputation systems on the platform, and user reviews. The fundamental objective of this approach is to circumvent the formidable challenge of standardizing and digitizing intrinsic quality attributes, such as flavor, texture, and freshness. These qualities are widely regarded as technically difficult and economically costly to standardize and digitize.

  While this approach establishes a trust framework through peripheral attributes, it overlooks the core flaw in the quality dimension. Consumer concerns primarily lie not in the product's provenance (e.g., its organic status), but rather in its intrinsic quality attributes (e.g., flavor and texture).

  \item \textbf{Direct Quality Assessment Based on Ontological Properties}
  
  This approach confronts the fundamental issue directly by employing an intelligent tiered algorithm to replicate offline quality inspection processes. However, the field as a whole succumbs to a methodological illusion: the pursuit of universal, absolute grading standards and the positioning of algorithms as the ultimate arbiters. This preoccupation with metrical precision obscures the commercial essence of trust-building, which is predicated on human value judgments.
\end{itemize}

The inherent limitations of these approaches are rooted in the fundamental contradictions inherent to the process of agricultural product grading, which is delineated by the so-called ``impossible triangle''.

\begin{enumerate}
  \item \textbf{Perishability}: Certain fruits and vegetables are characterized by a high degree of perishability, with a significantly limited shelf life. This necessitates that the entire process—from the initial picking and grading to the distribution stage—be executed with optimal efficiency and expediency, with the entire operation being completed within a reasonably brief timeframe. The efficacy of complex, sophisticated quality assessment techniques is rendered moot when products demonstrate signs of deterioration. This hinders the ability to effectively adapt to real-world application scenarios.
  
  \item \textbf{Biological Nature}: Agricultural products are naturally grown organisms, exhibiting significant variations in appearance, size, and internal quality among individual specimens. This high degree of variability poses significant challenges in establishing a single, universal, absolute quality grading standard, thereby increasing the complexity of standardized information collection. The presence of differences in regions, individuals, and even consumer preferences gives rise to divergent evaluations, thereby further complicating the establishment of a unified standard for addressing this issue.
  
  \item \textbf{Cost-effectiveness}: As staple consumer goods, most fruits and vegetables have low per-unit costs and narrow profit margins. Consequently, the financial implications of any quality inspection technology must be meticulously regulated. For a limited number of products with elevated premiums, while supplementary expenses may be accommodated to a certain extent, fundamental oversight of particular specimens remains imperative.
\end{enumerate}
This triangle fundamentally demonstrates the impracticality of an ``absolute grading standard,'' making it impossible to achieve high precision, high speed, and low cost simultaneously. This paper posits that addressing the trust deficit necessitates a fundamental paradigm shift, namely, the reorientation of the goal of hierarchical technologies from replacing human judgment to supporting human decision-making. Algorithms should not function as arbiters; rather, they should provide transparent, explainable decision-making foundations. Consequently, tiered information should serve as one of the bases for decision-making, functioning as guiding data rather than decision-making data. In order to illustrate this transformation, the present paper makes the following contributions:

\begin{itemize}
  \item \textbf{Theoretical Modeling}: The present study proposes and validates the Trust Pyramid Model, thereby confirming that the attributes of the quality layer form the foundation of online trust deficits. The prevailing issue in the online realm at present pertains to the inversion of this conventional paradigm. This paper introduces the Triangular Trust Index (TTI), a novel metric for assessing trust.
  
  \item \textbf{Technological Innovation}: The TriAlignXA framework is designed to address triangular constraints, with three core engines that directly tackle these issues.
  
  The first component is the Biological Adaptability Engine, which is comprised of the Feature Separation and Feature Surface functions. The purpose of the Biological Adaptability Engine is to unlock biodiversity.
  
  The second component is the Timeliness and Efficiency Engine (Recursive Inference/Process Weighting), which guarantees the punctuality of its operations.
  
  The third component is the Economic Optimization Engine (Model Repository/Transfer Learning), which is designed to ensure cost-effectiveness.
  
  \item \textbf{Trust Bridge}: A pre-mapping mechanism is introduced that encodes multidimensional process data (e.g., Gauss distribution and feature scores) into QR codes. The purpose of this mechanism is to convey quality evidence to consumers.
  
  \item \textbf{Data and Validation}: The Fruit3 dataset was released, and experiments demonstrated that TriAlignXA achieved an 85.87\% classification accuracy and outperformed baseline models under triangular constraints.
\end{itemize}

\section{Related Work: Two Pathways of Indirect Verification and Direct Assessment}
The persistent trust deficit in the online fruit and vegetable e-commerce sector has spurred extensive research, which can be categorized into two interrelated approaches: indirect Trust Verification Based on Additional Attributes and direct Quality Assessment Based on Ontological Properties of the products themselves. This section will critically analyze these two approaches, revealing their fundamental limitations in addressing the core challenge of building trust relationships with consumers.

\subsection{\textbf{Indirect Trust Verification Based on Additional Attributes}}

The initial approach, reliability validation, focuses on circumventing the substantial challenge of directly quantifying the intrinsic and perceived quality attributes of agricultural products, including flavor, texture, and freshness. Instead, it endeavors to establish trust indirectly through a series of peripheral, supplementary attributes. The primary mechanisms that have been identified through research and practical application in this field are as follows: systematic trust assurance mechanisms, socialized trust transmission mechanisms, and risk-hedging trust assurance mechanisms.

\subsubsection{\textbf{Systematic Trust Assurance Mechanism}}
This mechanism relies on institutional rules and technical systems to establish a transaction security network, with typical examples including: In examining the impact of social institutions on trust in e-commerce, Lu et al.\cite{5} adopted an institutional framework and sociocultural perspective. Through the implementation of simulation experiments, the study demonstrated the role of institutional factors in shaping market trust, thereby establishing the theoretical foundation for a systematic trust assurance mechanism. Wang et al.\cite{6}primary focus is on the technical design of margin policies, with a demonstration of how economic constraint mechanisms enhance trust levels by regulating trading behavior. Merhi\cite{7} expanded the research scope to encompass the national policy level, revealing that policy optimization can systematically mitigate transaction risks through macroeconomic regulation. The three aforementioned studies form a safeguard system for institutional trust.

Guo et al.\cite{8} identified the factors influencing online purchases of fruits and vegetables through factor analysis and regression analysis, including health consciousness, reputation, price, and convenience. These factors offer insights into consumer decision-making processes and demonstrate that optimizing these dimensions can address trust issues in this market.

Dong \cite{9} big data analysis further indicates that conventional technical rules, such as green certification, have limited effectiveness in fresh produce scenarios. This highlights the unique role of systematic safeguards in dynamic quality management.

The approach's limitations are evident in its inability to ensure the intrinsic quality of the product, a crucial aspect of consumer satisfaction. Consequently, it fails to address consumers' fundamental concerns regarding product quality.

\subsubsection{\textbf{Socialized Trust Transmission Mechanisms}}
This mechanism underscores the generation and transmission of trust through social relationships, group interactions, and information sharing.

In addressing the mechanisms through which operational platforms influence trust, Hajli\cite{10} employed Iranian social commerce data to elucidate the pivotal role of institutional trust in information sharing. Their research corroborates the notion that platform attributes—such as interface design and community norms—underpin the transmission mechanisms of social trust.Alkhalifah\cite{11} employed the PLS-SEM method to validate the direct driving effect of platform trust, while  Al-kfairy et al.\cite{12} discovered in the Instagram shopping context that trust propensity and online experience form a synergistic effect through platform attributes. The findings, when considered collectively, provide a comprehensive overview of the multidimensional pathways through which socialized trust is transmitted.

In the fresh food e-commerce sector, He et al.\cite{13} were the first to reveal the failure of traditional quality signals (such as packaging certifications), exposing the limitations of systematic safeguards in the fresh produce domain.Campos et al.\cite{14}identified cultural sensitivity to food shape anomalies through meta-analysis, a finding that poses unique challenges to the socialized trust transmission mechanism.

The distinctive characteristics of fresh food e-commerce further accentuate the limitations of social evaluation mechanisms: Hua's research revealed that consumers demonstrated an exceptionally low tolerance for abnormal fruit and vegetable appearances. Campos's findings confirmed that cultural biases distort the objectivity of social evaluations.

The fundamental limitation of this approach is predicated on the premise that it fosters trust through the subjective evaluations and group consensus of others, as opposed to direct assessments of objective quality. This renders it vulnerable to a variety of threats, including, but not limited to, fake orders, false reviews, information overload, and group bias. Consequently, it can be considered a fragile and potentially manipulable source of trust.

\subsubsection{\textbf{Risk-Hedging Trust Assurance Mechanisms}}
This mechanism has been demonstrated to influence consumer behavior by mitigating their perceived risk and offering post-purchase compensation. To illustrate, this mechanism mitigates risk perception through ex post compensation; however, it does not resolve the fundamental contradiction.

The impact of merchant service quality on trust exhibits risk-hedging characteristics.Senali et al.\cite{15} confirmed through Instagram data that comment quality directly impacts seller trust as a key risk buffer factor, highlighting the buffering role of UGC in mitigating information asymmetry. Zhao et al.\cite{16} after-sales guarantee strategy and Wu et al.\cite{17} green certification research, respectively, establish risk hedging mechanisms from the perspectives of post-purchase compensation and pre-purchase information transparency. These strategies are characterized by a common feature: the alleviation of consumers' quality concerns through either remedial measures after the fact or preemptive commitments.

In the domain of fresh produce, conventional risk management strategies may encounter challenges in terms of their efficacy. Senal's research revealed that post-sale policies frequently fall short in addressing the immediate financial setbacks associated with perishable goods. In addition, scholars have explored the potential of enhancing label information to bolster consumer trust, mitigate information asymmetry, and empower decision-making\cite{18}. Zhao's compensation mechanism is characterized by elevated implementation costs in the context of short-term insurance scenarios.

The approach's limitations are evident in its function as a financial substitute for ``trust'' based on economic compensation. Fundamentally, it functions as a financial instrument to mitigate risk associated with quality uncertainty rather than as an indication of quality itself. This document acknowledges the dearth of transparency surrounding quality information; however, it does not commit to addressing this issue.

\subsubsection{\textbf{Comprehensive Limitations: The Dilemma of Indirect Pathways}}
In summary, indirect trust verification pathways based on additional attributes—despite their diverse manifestations (systematic, social, risk-hedging)—all share a fundamental strategic flaw: they evade the digitalization and standardization of a product's most core ontological quality attributes.

While the aforementioned mechanisms have proven effective in general e-commerce, their application to the marketing of fruits and vegetables exposes a fundamental flaw: the non-transferability of quality signals. The blockchain technology has the capacity to verify the origin of a product; however, it does not possess the capability to convey the quality of the product's taste. User reviews, while they may accurately reflect a product's texture, are subject to emotional bias, as evidenced by statements such as, ``This apple is crisp, but not as sweet as expected.'' The subject exhibits an insufficient degree of adaptability with regard to the production of characteristics. Socialized mechanisms are often ineffective in addressing biological heterogeneity, as evidenced by the inability to account for variations in fruit taste among fruits from the same tree. Similarly, risk hedging mechanisms often fail to cover the immediate risks associated with short-shelf-life products, such as spoilage caused by delivery delays.

The theoretical deficit inherent in this approach is predicated on the observation that extant research has been chiefly oriented towards the study of durable goods, including electronics and daily necessities. This has resulted in a neglect of the dynamic quality characteristics exhibited by fresh agricultural products. This oversight renders the indirect trust mechanism ineffective in the context of fruits and vegetables. It is imperative that systematic mechanisms prioritize dynamic traceability, exemplified by real-time temperature control data, over static origin labeling. The integration of socialization mechanisms with AI-driven quality prediction models (e.g., predicting taste based on appearance) is imperative. The integration of socialization mechanisms with AI-driven quality prediction models (e.g., predicting taste based on appearance) is imperative.

The fundamental principle underlying indirect trust verification pathways is the substitution of sensory perception with institutional mechanisms. However, in the context of the fruit and vegetable sector, this approach must evolve to incorporate ``supplementing sensory perception with institutional mechanisms.'' This shift provides the theoretical foundation for the TriAlignXA framework proposed in this study.

\subsection{Direct Quality Assessment Based on Ontological Properties}
In contrast to the ``detour'' strategy of indirect pathways, the direct quality assessment approach based on ontological attributes confronts the core challenge head-on. The objective of this study is to quantify and evaluate the physical, chemical, and sensory properties of agricultural products through technical means. The aim is to replicate or even surpass the quality grading capabilities of human experts. The evolution of this approach is consistent with a discernible trajectory of technological advancement, beginning with metric-based correlation and progressing to image-based automatic classification. Ultimately, the approach is directed toward multimodal integrated evaluation.

\subsubsection{\textbf{Rule Definition: Indicator Association Construction}}
Preliminary research was chiefly predicated on the ``indicator-based association'' paradigm. The fundamental premise of this approach is as follows: First, artificially designed features that are highly correlated with quality (e.g., fruit diameter, color histogram, texture contrast, etc.) are to be extracted manually through image processing techniques. Subsequently, machine learning models should be employed to establish statistical correlation models between these visual indicators and internal quality or defect labels.

For instance, Blasco et al.\cite{19} achieved breakthroughs in defect detection and size estimation through Bayesian discriminant analysis. Dhakshina Kumar et al.\cite{20} employed microcontroller-based imaging technology to facilitate precise grading of tomatoes, while Chithra et al.\cite{21} research concentrated on the fine segmentation of apple images to enhance the accuracy of defect recognition. Varghese et al.\cite{22}developed a vision-based system for shelf-life prediction. Nirale et al.\cite{23} applied this approach to growth monitoring, and Costa et al.\cite{24} even attempted to indirectly determine chemical constituents, such as total phenolic content, through image colorimetry.

This paradigm successfully demonstrated the feasibility of machine vision in hierarchical tasks, thereby laying the foundation for the field. However, the process is dependent on expert knowledge for feature design and selection, which can lead to cumbersome procedures and suboptimal generalization capabilities. Of greater significance is the methodological limitation concerning the assumption that a stable, quantifiable mapping relationship exists between appearance metrics and intrinsic quality. This assumption is frequently found to be tenuous and one-sided, particularly in the context of agricultural products characterized by remarkably elevated biodiversity. This phenomenon results in an inability to encapsulate the intricacies and non-linear qualities inherent in these products, consequently leading to a deficiency in the model's stability.

\subsubsection{\textbf{Vision for Automation: Image Classification Tasks}}
The advent of deep learning has precipitated a paradigm shift in research methodologies, with a concomitant focus on ``image-based automatic classification.'' Convolutional neural networks (CNNs) have the capacity to automatically learn hierarchical features from pixels, thereby eliminating the necessity for arduous manual feature engineering. The objective of this initiative is to transform grading tasks into end-to-end image classification problems, thereby achieving fully automated grading.

Nikhitha et al.\cite{25}achieved accurate disease classification using CNNs with transfer learning. Saranya et al. \cite{26} study yielded a categorization of bananas into four maturity classes, a finding that has significant implications for the banana industry.Khamis et al.\cite{27} integrated ResNet50 with YOLOv3 to assess the maturity of palm oil fruit bunches, while Kuo et al.\cite{28} employed neural networks to identify defects in apples. Morshed et al.\cite{29} applied DenseNets, while Kale and Shitole\cite{30} enhanced pomegranate grading accuracy through optimizer research. Gokhale et al.\cite{31} utilized the ConvNeXt model to assess tomato freshness.

This paradigm has advanced the automation of classification processes significantly, achieving remarkable accuracy on specific datasets. However, it collectively fell into another methodological illusion: the pursuit of a universal model capable of serving as the ultimate ``arbiter.'' This approach places excessive reliance on data-driven methodologies, neglecting the inherent ``impossible triangle'' constraints inherent to agricultural product grading. Consequently, models frequently demonstrate high performance on closed datasets; however, their robustness and practical deployment value are significantly diminished when confronted with issues such as inconsistent grading standards, highly subjective labeling, and high computational costs. The text provides an answer to the question, ``What is the level?'' However, it does not provide an explanation for why the level is as it is. Consequently, the credibility and practicality of the aforementioned approach are limited.

\subsubsection{\textbf{Multimodal Feature Fusion and Ensemble Evaluation}}
The following section is concerned with the integration of features. Multimodal Ensemble Evaluation. Acknowledging the constraints of the prevailing paradigm, research initiatives have commenced exploration of the ``multimodal integrated evaluation'' approach. This approach endeavors to overcome the constraints imposed by a singular visual modality by integrating multi-source information from optical, acoustic, tactile, and even chemical sensors. The objective is to establish a more comprehensive and robust quality assessment system.

Su et al.\cite{32} employs advanced image processing algorithms to integrate features, thereby enabling high-accuracy grading of potatoes. Ren et al. \cite{33} innovative approach entailed the integration of multiple machine learning models, including support vector machines (SVM), K-nearest neighbors (KNN), and decision trees, to conduct a comprehensive evaluation of apples and mangoes. Bhargava's\cite{34,35} research examined multi-classifier fusion and automatic detection algorithms. In his work, Mohapatra et al.\cite{36} employed CNN (convolutional neural network) and R-CNN (region-based convolutional neural network) techniques. Gurubelli et al.\cite{37}integrated texture and color features to achieve fine-grained classification. In his study, Wagimin et al.\cite{38} integrated support vector machine (SVM), artificial neural network (ANN), and k-nearest neighbor (K-NN) algorithms to classify mangoes based on physical parameters. Similarly, Shi et al.\cite{39}combined multi-view spatial networks with bidirectional long short-term memory (Bi-LSTM) to achieve exceptionally high accuracy in apple classification.

Multimodal approaches signify a substantial advancement toward a more comprehensive perception, acknowledging the multidimensional nature of quality. Nevertheless, extant research continues to exhibit two fundamental deficiencies: First, it remains fundamentally a ``feature fusion'' technique, failing to resolve the inherent contradiction of ``algorithms as arbiters'' at a paradigmatic level. Furthermore, the acquisition and processing of multimodal data serves to compound the trade-off between timeliness and cost efficiency within the ``impossible triangle.'' Secondly, and more crucially, the majority of these studies remain confined within a ``vision-dominant'' paradigm. The so-called multimodal approaches frequently consist of merely combining different visual features, thereby failing to genuinely integrate heterogeneous sensory information such as weight, hardness, and sugar content. Consequently, their classification results are inadequate in comprehensively reflecting the end-user's overall experience, thus failing to achieve a true paradigm shift.

\subsubsection{\textbf{Comprehensive Limitations: The Dilemma of Direct Pathways}}
In summary, while the direct evaluation approach based on ontological attributes continues to evolve technologically, its development remains constrained by a shared vision: the pursuit of a universal, objective, and automated absolute grading standard, with algorithmic models serving as the ultimate arbiters of this standard. This vision disregards the ``impossible triangle'' concept, which is a theoretical framework representing the interwoven challenges of the inherent biodiversity of agricultural products, the stringent demands for processing efficiency, and the limited financial resources available to the industry. Consequently, irrespective of the paradigm, attaining an optimal balance between accuracy, efficiency, and cost remains a formidable challenge.

Existing approaches have been unsuccessful in achieving their primary objective of fostering trust due to their failure to recognize the commercial essence of trust. The commercial essence of trust can be defined as the support of consumers' subjective value judgments rather than the replacement of these judgments. Consequently, a paradigm shift is imperative to reorient technological objectives from the pursuit of ``absolute classification'' toward the provision of ``transparent and explainable decision-making foundations.'' This reorientation will systematically resolve the ``impossible triangle'' dilemma. The TriAlignXA framework, which is the subject of this paper, is precisely the product of this paradigm shift.

A critical review of extant research on the necessity of paradigm shifts reveals that, despite divergent approaches, both indirect validation and direct assessment methods have fundamentally failed to resolve the challenge of trust in online fruit and vegetable transactions. This failure does not stem from the inherent shortcomings of the technology itself, but rather from its shared flawed premise: that trust can be established through substitute proofs or absolute adjudication.

Indirect verification pathways endeavor to supplant peripheral, derivative evidence—such as origin, reviews, and after-sales policies—with direct recognition of the product's intrinsic quality. The system establishes a ``safety net'' for transactions, yet it has been observed that it leaves a significant ``black hole of quality information.''

The direct evaluation approach endeavors to ascertain an absolute, universal quality rating through the utilization of algorithmic models. Despite its rigorous confrontation with quality, the field has fallen prey to methodological illusions and the practical dilemma of the ``impossible triangle'' by pursuing unrealistic ``objective standards.''

A common limitation of both approaches lies in their attempt to bypass or replace the central role of humans (consumers) in quality assessment and their value judgments. Nevertheless, the fundamental principle of trust does not originate from the precision of algorithms or the intricacy of systems. Instead, it is derived from consumers' subjective confidence that product quality is commensurate with their expectations. This confidence cannot be gained through ``avoidance'' or ``arbitration,'' but only through ``empowerment'' and ``support.'' Consequently, the pivotal to surmounting the trust deficit is not to perpetuate the remediation of extant pathways, but rather to effect a foundational paradigm shift: The core objective of technology must shift from serving as the ultimate ``arbiter'' to providing a transparent ``decision-making basis.''

This signifies a shift in the design philosophy of technological solutions: a transition from the objective of replacing human judgment to the focus on supporting human decision-making; and a transition from delivering a closed, hierarchical conclusion to providing open, explainable quality evidence. The classification information should serve as one of the bases for decision-making, functioning as guiding data rather than decision-making data. The evaluation of its success should not be constrained by improvements in algorithmic metrics alone. Instead, its capacity to effectively translate offline quality perceptions into credible online information should be considered. This ability empowers consumers to make confident purchasing decisions.

This paradigm shift constitutes the theoretical foundation of the TriAlignXA framework proposed in this study. In order to achieve this transformation in a systematic manner, it is necessary to deconstruct the following elements: The following inquiry is posited: How is such value assessment effectively executed in offline scenarios? This inquiry will serve as the foundational framework for the subsequent chapter.

$\,$

$\,$

\section{The Trust Dilemma: An Exploration of the Dilemma Surrounding the Grading of Fruit and Vegetables}
This chapter explores trust-building mechanisms in the context of fruit and vegetable e-commerce, with a focus on addressing user needs. The present study draws on established offline models and analyzes their market processes to examine the multidimensional factors of trust. It proposes a "Trust Pyramid" model. Subsequently, it innovatively introduces a "dual-source data" perspective, analyzing the fundamental challenges this model faces during online migration from two angles: user needs and the coupling between user demands and platform supply capabilities. This finding serves to identify the fundamental problem that this research seeks to address.

\subsection{Trust Pyramid Model}

The offline fruit and vegetable sales chain operates within a relatively transparent ecosystem where trust is established through a coherent process. Products undergo a series of steps, including harvesting, sorting, and grading, before entering market circulation. Retailers establish an environment replete with information through product display, origin labeling, transparent operations, and brand building, thereby providing consumers with sufficient basis for purchasing decisions. In this process, grading primarily serves as an initial screening mechanism—one component within the multifaceted trust-building framework of offline retail, not its entirety.

\begin{figure}[htbp]
\centering
\includegraphics[width=1.0\textwidth]{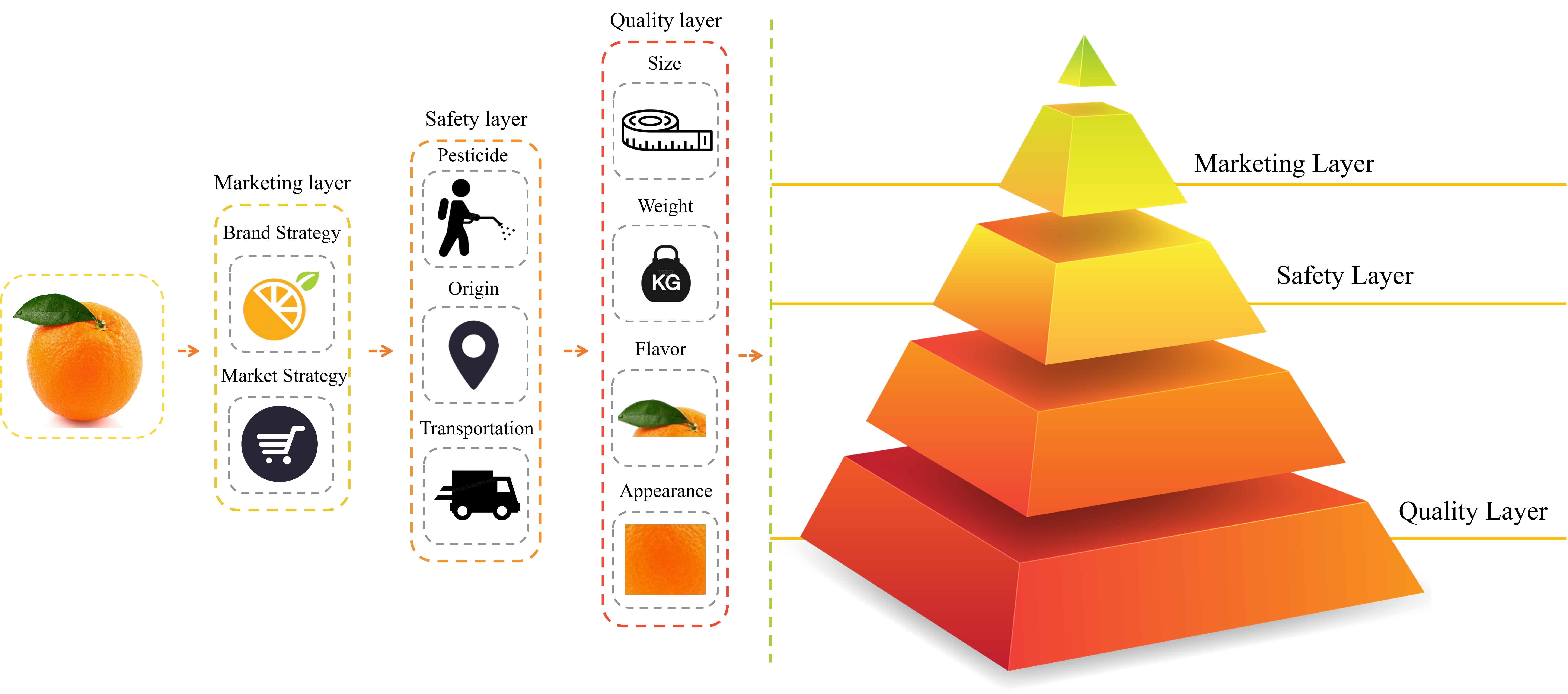}
\caption{Schematic Diagram of the Trust Pyramid.Left: illustration of corresponding grading metrics (from left to right); Right: the layered model of the Trust Pyramid. The model contains more than three layers to reflect important subdivisions even within a single level.}
\label{fig:image1}
\end{figure}

The present paper proposes a Trust Pyramid Model that integrates the complex factors influencing consumer trust into three core tiers: the Quality layer, the Safety  layer, and the Marketing layer. This model is illustrated in Figure \ref{fig:image1}.

\begin{itemize}
  \item \textbf{Quality layer:}
The quality layer is concerned with the physical attributes of fruit and vegetable products, including appearance, texture, flavor, and nutritional value. These attributes are directly linked to consumers' expectations and evaluations of product quality, forming the most immediate and fundamental foundation of trust.

  \item \textbf{Safety layer:}
This layer reflects consumers' assessment and avoidance of food safety risks, serving as an indispensable safeguard in building trust.The safety layer comprises a compendium of information regarding the product, including its provenance, the management practices employed in its cultivation, and the certifications attesting to its food safety.

 \item \textbf{Marketing layer:}
The marketing layer is comprised of intangible factors, including brand image, pricing strategies, and after-sales service.The influence of emotions on consumer perceptions and purchase motivations is a critical factor in the indirect impact on purchasing decisions, thereby serving as a pivotal stage for the development and reinforcement of trust.
\end{itemize}

The model posits that the three tiers do not contribute equally to consumer trust; rather, they form a pyramid-like hierarchical structure. The quality layer serves as the foundation, the safety layer provides support, and together they build the bedrock of trust. The marketing layer functions as the value-added structure at the apex. Lower layer contain elements that are more closely associated with the product's intrinsic attributes, representing fundamental prerequisites and the foundation for establishing trust. Higher layer exhibit a greater propensity to prioritize external value-added attributes, which play a pivotal role in reinforcing trust and facilitating decision-making processes. The validity of this theoretical model has been confirmed through large-scale consumer research and statistical validation (detailed results are presented in the experimental section of Chapter 5).

\subsubsection{\textbf{Quality Layer: The Foundational Entity of Trust}}

In the analysis of the myriad factors that influence consumer purchasing decisions, the assessment of product quality emerges as the most direct and fundamental element.This initial stage of product quality assessment exerts a significant influence on consumers' final decisions.A thorough examination of its underlying logic and principles is warranted.A thorough examination of its underlying logic and principles is warranted. Consumers assess the quality of fruits and vegetables through a multifaceted evaluation process that incorporates visual inspection, encompassing attributes such as color, shape, and integrity, complemented by direct sensory experiences including texture, freshness, and flavor.Furthermore, the nutritional value of the food is a significant consideration.In this process, the visual quality of fruits and vegetables is closely linked to their intrinsic attributes, which collectively form the primary basis for consumer purchasing decisions\cite{40,41,42}.These evaluation metrics closely align with consumer expectations, directly determining whether a product meets anticipated standards.Consequently, the quality hierarchy serves as the foundational cornerstone for building consumer trust.

The evolution of grading systems has been inextricably linked to technological advancements. Early grading systems relied on manual or mechanical physical indicators, such as size, weight, and defects, for classification purposes. The advent of artificial intelligence has precipitated a paradigm shift in conventional grading methodologies, thereby inaugurating a contemporary epoch characterized by elevated efficiency and precision\cite{43,44,45,46}.

\subsubsection{\textbf{Safety Layer: Trust Assurance Framework}}

The safety layer, which is predicated on the foundation of quality, provides consumers with psychological assurance against physical and chemical hazards, thereby addressing concerns about whether a product is harmless.Information at this level is typically not directly accessible through sensory perception, making it highly dependent on credible signaling.

The heightened awareness among consumers regarding food safety concerns has emerged as a pivotal factor influencing their purchasing decisions\cite{47,48}.This phenomenon is intricately intertwined with the pressing reality of frequent foodborne illnesses associated with fresh agricultural products, a problem that is further exacerbated by the presence of pesticide residues resulting from excessive fertilizer and pesticide use\cite{49,50}.Moreover, mechanical damage incurred during harvesting and transportation has been substantiated as a probable food safety concern\cite{51,52}.

The security layer functions as a protective shield for the quality layer.Even if a product is of the highest quality, trust cannot be established if its security is questionable.

\subsubsection{\textbf{Marketing Layer: The Value-Added Potential of Trust}}

The marketing layer, situated at the apex of the marketing pyramid, has been shown to activate emotional resonance, reduce decision friction, and ultimately achieve value enhancement.The efficacy of this system is contingent upon the establishment of a robust foundation comprised of quality and security layers.

Brand Value: A brand signifies the ongoing accumulation of quality and safety commitments, thereby functioning as a credit pledge that connects past promises with future expectations.It endows products with emotional value and cultural significance that transcend their physical attributes\cite{53,54}.

Pricing and Promotion Strategy: The term "premium pricing" is often associated with high quality, while "promotional pricing" is known to directly stimulate impulse purchases.Price functions as the most direct regulator of perceived value.

Pursuant to the subject matter at hand, the following is a delineation of the terms of the after-sales service. Promises such as "money-back guarantee if not satisfied" represent a risk reversal strategy. By allocating a portion of the risk associated with the purchase decision from consumers to merchants, these platforms significantly mitigate consumers' perceived risk, thereby fostering an environment conducive to trust\cite{55}. 

The internal logic of the pyramid is as follows: These three tiers do not exist in isolation; rather, they follow a strict supporting logic. Absent a robust quality layer, the assurances provided by the security layer become ephemeral, akin to a castle in the air. Absent the presence of quality and security layers to provide support, investments in the marketing layer will inevitably yield diminishing returns, or worse, risk backfire by eroding trust through exaggerated claims.The value proposition of a product, which encompasses its functionality, emotional appeal, and perceived reliability, is determined by the quality layer and the marketing layer. These factors collectively influence consumers' propensity to incur cost in exchange for the product.The structural integrity and hierarchical relationships of this pyramid model have been rigorously validated through empirical research. The ensuing chapter contains detailed data and statistical analysis results.

\subsection{Challenges to Online Trust: inverted pyramid}

In an effort to cultivate a comparable level of confidence online, these e-merchants have employed a variety of strategies, including the implementation of origin certification at the security layer, the development of brand identity in the marketing layer, and the provision of after-sales guarantees. These measures establish a certain level of superficial trust. However, the fundamental deficiency of the online paradigm is the considerable absence of quality-layer information, resulting in the online trust pyramid losing its foundational basis. This results in an imbalanced structure that is top-heavy and unstable at its core.

In order to methodically analyze this issue, the present study proposes a dual-source data comparison approach.

\begin{itemize}
  \item \textbf{User Demand Importance Analysis:}
This phenomenon is indicative of consumers' fundamental, unadulterated preferences and needs.

  \item \textbf{The 7-Point Coupling Importance Score:}
It is a metric used to determine the relative importance of a given coupling. This phenomenon is indicative of the "perceived importance" that users ascribe to various categories of information, as influenced by the prevailing conditions of platform supply. This phenomenon can be conceptualized as the interplay between user demand and platform supply capacity, thereby offering insights into the user perspective.
\end{itemize}

A critical examination of the discrepancies and tensions between these two datasets can unveil the structural flaws in the current online trust system. Specifically, a considerable "coupling gap" has been identified, which refers to the disparity between the absolute demands of users and the supply capabilities of the platform. This discrepancy ultimately gives rise to an unstable "inverted pyramid" trust structure.

\begin{enumerate}
  \item \textbf{Quality Layer:} The advent of absolute demand, concomitant with the paucity of core supply, is a phenomenon of considerable pertinence.

The quality layer exerts a dominant influence on absolute demand data. A significant proportion of consumers, 81.71\%, regard quality assurance as the paramount guarantee. Moreover, an overwhelming 88.33\% prioritize "product appearance" as their predominant decision factor, underscoring the paramount importance of quality information.

However, in the coupled-attribute ratings, "freshness" (5.84) and "taste/flavor" (5.72) received the highest importance scores. This phenomenon underscores the current platforms' most significant vulnerability, as the top-rated demands pertain to the sensory experiences that are most challenging to convey in online transactions. The high scores in the quality layer are not an indication of the platform's strengths; rather, they serve as a strong indictment of its significant core capability gaps. These gaps have exposed a substantial absence of reliable information at the foundation of the online trust pyramid. The high emphasis placed on these needs by users stems precisely from their profound sense of unmet needs.
  
  \item \textbf{Safety Layer:} The Paradox of Absolute Need in Relation to Perceived Importance.

An examination of the data from the safety layer reveals a critical paradox. In absolute need metrics, "pesticide residues" is the most significant factor, accounting for 83.66\% of the total. This indicates that the presence of pesticide residues is the fundamental and unmet need for users.

However, in the coupling scale ratings, "pesticide residue testing" (4.98) is perceived as relatively unimportant. This discrepancy serves to elucidate the "coupling effect": in the absence of transparent quality-layer information to support it, isolated safety claims (e.g., an unexplained certification certificate) are susceptible to losing credibility among users. It is evident that users do not exhibit a state of indifference regarding safety concerns; rather, they harbor a profound inability to place complete trust in the safety information provided by the prevailing platform, a circumstance that is predicated upon its disconnection from their quality experience. Consequently, its perceived importance is diminished, reflecting the significant failure of the safety layer to function as a "pillar of trust" under the current model.
  
  \item \textbf{Marketing Layer:}The advent of costly substitutes and the subsequent accumulation of systemic risks are phenomena of increasing concern.

The coupled data (scale scores) from the marketing layer reveal the current state of affairs. Platforms are compelled to allocate resources toward dimensions such as "after-sales service" (5.58) and "user reviews" (5.56). These high-cost, actionable "additional proofs" are used to attempt to plug massive gaps in quality and safety.

While this strategy may appear to be effective in the short term, it ultimately results in a top-heavy inverted pyramid. It has been demonstrated that high scores at the marketing layer are indicative of platforms compensating for deficiencies in their core capabilities through the implementation of costly substitute solutions. Fundamentally, this process entails the transfer of risk (e.g., post-sales compensation guarantees) and collective endorsement (e.g., user reviews). However, it neglects to tackle the crucial issue of inadequate quality information. In the event of major public relations crises or post-sale disputes, the trust that has been established can rapidly disintegrate, resulting in substantial damage to the brand. The resulting costs far exceed the initial investment in quality digital infrastructure. This phenomenon signifies the most substantial systemic risk inherent within the prevailing trust framework.

In summary, the dual-source data perspective provides a comprehensive overview of the current predicament. However, coupled demands (scale rating data) reveal that platform resources and user perception are distorted toward marketing and after-sales. This distortion unveils the unsustainability of the prevailing model and offers the most compelling rationale for the paradigm shift proposed in this paper.
\end{enumerate}

\subsection{The Triple Dilemma of Fruits and Vegetables}

\subsubsection{\textbf{Impossible Trinity}}
\begin{figure}[htbp]
\centering
\includegraphics[width=0.5\textwidth]{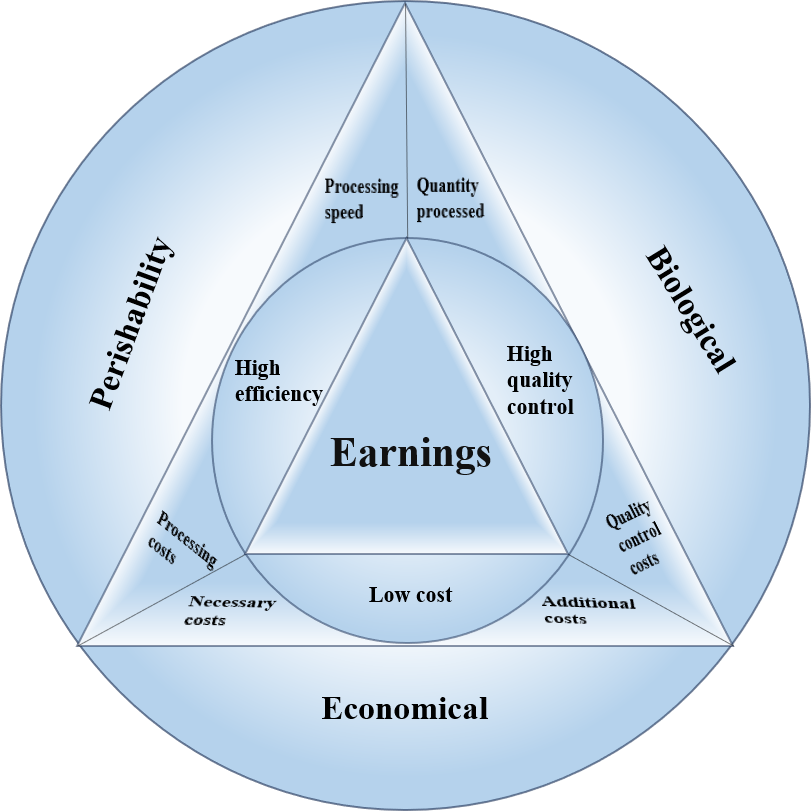}
\caption{Schematic Diagram of the Impossible Triangle: The outer layer represents the characteristics of fruits and vegetables; the intersection of two conflicting features within the triangular area indicates inherent contradictions; the central circle corresponds to the essential requirements; and among these three essential requirements, the most critical one constitutes the fundamental proposition.}
\label{fig:image2}
\end{figure}

A thorough examination of the available data reveals that fruit and vegetable products possess unique characteristics that render this model inapplicable.Consequently, in order to effectively address issues of trust in online fruit and vegetable sales, it is essential to refine the self-certification mechanism for quality.Consequently, efforts should also focus on these inherent characteristics.As demonstrated in Figure \ref{fig:image2}, this predicament emanates from the interrelated constraints of three fundamental attributes: biological nature, time-sensitivity, and economic factors. These attributes collectively render the standardization and conveyance of quality information arduous.

\begin{enumerate}
  \item \textbf{Perishability:} Certain fruits and vegetables are highly perishable, with extremely limited shelf lives.This necessitates a relatively efficient and rapid workflow throughout the entire process—from harvesting and grading to distribution—which must be completed within a reasonably short timeframe.The employment of sophisticated, thorough, and time-consuming quality assessment techniques is rendered obsolete due to product spoilage.This characteristic renders them challenging to adapt to real-world application scenarios.
  
  \item \textbf{Biological Nature:} Agricultural products are naturally grown organisms that exhibit significant variations in appearance, size, and internal quality among individuals.This high degree of variability makes establishing a single, universal, absolute quality grading standard extremely difficult, greatly increasing the complexity of standardized information collection.The establishment of a unified standard is further complicated by differences in geographical regions, individual characteristics, and even consumer preferences, which lead to divergent evaluations.

  \item \textbf{Cost-Effectiveness:}As daily consumer goods, the unit prices and profit margins of most fruits and vegetables are relatively limited.Consequently, the financial implications of any quality inspection technology must be meticulously regulated.For a limited number of high-premium products, while supplementary expenses may be reasonably mitigated, fundamental oversight of particular specimens remains imperative.
\end{enumerate}

These three attributes are inextricably linked, forming a fundamental constraint that shapes the system's behavior.

\begin{itemize}
  \item \textbf{}
The pursuit of high-quality standards, in response to concerns regarding biodiversity, frequently necessitates the implementation of more complex algorithms and extended processing times. This can result in a compromise of timeliness and an increase in costs, potentially undermining the economic viability of the process.

  \item \textbf{}
The pursuit of extreme timeliness has been demonstrated to impose limitations on the complexity of available technologies, thereby creating significant challenges in addressing the issues associated with biodiversity.

 \item \textbf{}
Stringent economic constraints simultaneously preclude the implementation of high-cost technical solutions that could effectively address biodiversity concerns and ensure timeliness.
\end{itemize}

\subsubsection{\textbf{New Evaluation Indicator: Triangular Trust Index (TTI)}}

In order to formally describe this fundamental constraint, this paper constructs a conceptual model and defines a new evaluation metric—the Triangular Trust Index (TTI). The following concepts form the foundation of this metric:

\begin{itemize}
  \item \textbf{Information Coverage Quality (ICQ) index}
The definition of a variable is to be established as the Information Coverage Quality (ICQ) index, which is defined as the ratio of the actual system's provided confidence coverage interval (scq) to the user's required confidence coverage interval (ccq).

This metric is indicative of the extent to which the model fulfills user trust requirements, with 1 representing the pass threshold and 2 denoting the highest attainable level. The specific numerical value is calculated based on the acquisition definitions of scq and ccq.The metrics known as SQ and CCQ are both measured in Trust Factors (T), which serve to quantify the importance of a given metric in information presentation.Each dimension possesses its own unique metrics. For instance, one-dimensional classification guidance information can assist consumers in conducting preliminary screening to a certain extent. While other detailed requirements can be adjusted based on individual preferences, including parameters, weight information, size details, two-dimensional image data, video content, and three-dimensional modeling information. Each metric across these dimensions corresponds to a distinct set of trust factors. These trust factors are recorded in a regional classification dictionary and undergo dynamic updates. The weighting of each evaluation metric within the specific tiers of the trust pyramid is derived from the metric's weighting factor. The tier weighting system is structured according to a hierarchical framework, wherein lower tiers are assigned a higher relative importance. Within a given tier, the relative importance of a factor corresponds to its trust weight.The trust pyramid itself is a process of continuously updating indicators, thus involving dynamic calculations. In this context, under the assumption that the trust factor $T_{e}$ is necessary for a feature e at layer i, the trust weight at that layer is designated as $s_{i}$ ($s_{i}$ 	
$>$ 0), while its inherent importance is denoted as $t_{e}$, as illustrated in Equation \eqref{eq:equ1}:
\begin{equation}
 {{T}_{e}}={{s}_{i}}\times {{t}_{e}}
\label{eq:equ1}
\end{equation}

The user-required trust level, ccq, is defined as the sum of trust factors deemed essential for specific user judgments. The essential trust factors, including weight and size, are maintained in a dynamic array $T_{need}$ with a total of n elements, as demonstrated in Equation \eqref{eq:equ2}:
\begin{equation}
ccq=\sum\limits_{1}^{n}{{{T}_{need}}_{(i)}}
\label{eq:equ2}
\end{equation}
In this system, scq denotes the trust factor  $T_{provided}$ by the system, with a total of m elements, as shown in Equation \eqref{eq:equ3}:
\begin{equation}
scp=\sum\limits_{1}^{m}{{{T}_{provide(i)}}}
\label{eq:equ3}
\end{equation}  
The ICQ calculation entails the determination of the minimum value between the excess supply gain and the ccq/scq value, thereby quantifying trust coverage quality.A parameter $\gamma$ is defined as the excess supply gain, which has the capacity to regulate ICQ in specific scenarios, thereby ensuring its maintenance within an acceptable range. The specific values are determined by the product and environment settings, which are defined in the regional classification metric dictionary as shown in Equation \eqref{eq:equ4}:
\begin{equation}
ICQ=\min (\frac{ccq}{scq},1+\gamma)
\label{eq:equ4}
\end{equation}

  \item \textbf{Fruit and Vegetable Processing Efficiency (FE)}
The variable is defined as the fruit and vegetable processing efficiency, or FE, which is expressed as the "ratio of system throughput to the spoilage rate threshold." In other words, FE is calculated as Actual Processing Throughput (AT) divided by Spoilage Rate Threshold (R), as depicted in Equation \eqref{eq:equ5}.

\begin{equation}
FE=\frac{AT}{R}
\label{eq:equ5}
\end{equation}

Among these: The decay rate threshold, R, is determined by the biological characteristics of the agricultural products in question. The decay rate threshold can be found in the regional data dictionary for reference.Throughput: The number of qualified samples that are processed per unit of time is measured in samples per minute.

 \item \textbf{Fruit Cost (FC)}
The term "FC (Fruit Cost)" is used to denote the proportion of economic cost. The economic cost, herein referred to as "C," is formally defined as follows: The term "FC" is defined as the total cost (TC) divided by the market price (P), as illustrated in Equation \eqref{eq:equ6}.

\begin{equation}
FC=\frac{TC}{{{P}_{market}}}
\label{eq:equ6}
\end{equation}

The cost of the system encompasses the expenses associated with its full lifecycle, including hardware depreciation, algorithm computation, energy consumption, and maintenance.The denominator encompasses the $P_{market}$, defined as the mean market price of the agricultural product, and $FC_{max}$, representing the maximum acceptable proportion. These values are stored in the regional data dictionary for reference.

 \item \textbf{Triangular Trust Index (TTI))}
The Triangular Trust Index (TTI) is defined as a function of three key variables: TTI can be expressed as a function of ICQ, FE, and FC. In this model, I(ICQ) is equivalent to credibility, E(FE) is analogous to processing efficiency, and C(FC) is equivalent to economic cost.

The inherent Perishability, Biological Nature, andCost-Effectiveness determine an intrinsic trade-off among these three variables, as demonstrated in Equation  \eqref{eq:equ7}:

\begin{equation}
TTI={{w}_{I}}\times ICQ+{{w}_{E}}\times FE+{{w}_{C}}\times (1-\frac{FC}{F{{C}_{\max }}}) 
\label{eq:equ7}
\end{equation}

The Weighting Allocation Principle stipulates that the sum of the weights of the I, E, and C categories must equal 1.The objective of maximizing weight-for-length ratio ($w_{I}$) is achieved. Given that the establishment of trust constitutes the fundamental objective of the system, this finding corresponds with the underlying conclusion that the "quality layer" functions as the foundational element within the Trust Pyramid model. It is recommended that $w_{I}$ be set to 0.6.

In the fresh produce system, ensuring timeliness and combating corruption are two fundamental constraints. While $w_{E}$ (FE weight) is secondary, it is nonetheless an important consideration.It is recommended that  $w_{E}$ be set to 0.3.

The objective of minimizing  $w_{C}$ (FC weight) has been achieved. The economic cost is a mandatory hard constraint, reflected in the penalty term. It is recommended that $w_{C}$ be set to 0.1.The following is a detailed explanation of the third term in the formula (1\textendash FC/$FC_{max}$):The term in question has been demonstrated to transform cost constraints into performance contributions. When FC is 0, this term assumes its maximum value of 1. When FC = $FC_{max}$, this term becomes 0, indicating that costs have reached the upper limit and contribute nothing to TTI.This phenomenon ensures that TTI increases as costs decrease.

The inherent biological nature, time-sensitive nature, and economic nature of agricultural products create an inherent trade-off among these three variables:

\begin{equation}
frac{\partial FC}{\partial ICQ}>0
\label{eq:equ8}
\end{equation}

The present study sought to examine the correlation between enhanced information coverage quality and elevated costs.As demonstrated in Equation  \eqref{eq:equ8}, to acquire more comprehensive and precise agricultural product information (e.g., high-definition imaging of individual apples, sugar content detection, and traceability to the growing farm), it is imperative to implement higher-performance sensing devices, more powerful computing units, and more complex data management systems.The technological requirements for enhancing information granularity and dimensionality typically entail significant capital and operational investments. These investments include procurement and maintenance costs for advanced sensing equipment, high-computing-power hardware, and efficient databases. Consequently, a direct positive correlation exists between "information quality" and "cost."As demonstrated in Equation \eqref{eq:equ8}, enhancing the ICQ is contingent upon increasing the feature dimension and measurement accuracy, a process that will result in a direct increase in the system's total cost.

\begin{equation}
frac{\partial FC}{\partial FE}>0
\label{eq:equ9}
\end{equation}

A positive correlation has been observed between the enhancement of processing efficiency and the augmentation of costs.As demonstrated in Equation \eqref{eq:equ9}, to enhance the operational efficiency of the sorting and packaging line—that is, to process a greater quantity of apples within a given time frame—it is imperative to incorporate advanced, high-efficiency automated equipment, such as accelerated robotic arms and enhanced conveyor systems with higher precision.Consequently, such efficiency gains are frequently accompanied by increased capital and operational expenditures, including equipment investments, power consumption, and maintenance costs.Consequently, in the pursuit of augmented processing capacity, a substantial positive correlation exists between "efficiency" and "cost": the implementation of more expeditious production cycles necessitates corresponding increases in capital investment.

\begin{equation}
frac{\partial TTI}{\partial FC}\le 0   (FC\to FC\_\max )
\label{eq:equ10}
\end{equation}

The convergence of diminishing marginal returns of cost inputs with performance is a critical factor in the analysis.As demonstrated in Equation  \eqref{eq:equ10}, during the initial phase, capital investment, e.g., the procurement of fundamental testing equipment, can considerably enhance product reliability and quality. At this stage, marginal benefits are high, and the investment yields clear returns.However, when costs have already reached a high level (e.g., when top-tier equipment is already in use), further investments aimed at achieving marginal performance gains (e.g., increasing equipment speed from 100 mph to 101 mph) yield diminishing returns, meaning the quality improvements generated become extremely minimal.Under these circumstances, continuing to increase capital investment is no longer an effective solution. Mathematically speaking, the rate of change (or marginal gain) in the corresponding function typically approaches zero, and in practice rarely takes on negative values.

\begin{equation}
frac{{{\partial }^{2}}TTI}{\partial ICQ\partial FE}<0
\label{eq:equ11}
\end{equation}

The Dilemma of Dual-Objective Optimization: Achieving an equilibrium between the precision of information and the velocity of its processing is imperative in contemporary information systems.As demonstrated in Equation  \eqref{eq:equ11}, when confronted with resource limitations (e.g., cost and time), the concurrent pursuit of both exceptionally high information quality (e.g., prolonging single-sample detection time to enhance accuracy) and remarkably high processing efficiency (e.g., managing a substantial number of samples per unit time) gives rise to an inherent conflict, thereby characterizing a quintessential multi-objective optimization dilemma.The simultaneous pursuit of these two objectives by the system may potentially compromise its stability and elevate error rates, thereby undermining the fundamental objectives it is designed to safeguard: namely, product reliability and overall quality.In many cases, a viable solution necessitates the balancing of trade-offs and optimization rather than the simultaneous maximization of all performance metrics.In many cases, a viable solution necessitates the balancing of trade-offs and optimization rather than the simultaneous maximization of all performance metrics.

In order to optimize the target metric TTI (product reliability and quality), three approaches must be considered: enhancing information and quality control (ICQ), improving processing efficiency (FE), and reducing the total cost of the system (FC).However, both of the first two strategies rely on the same resource—namely, the input of FC (cost)—and this input follows the law of diminishing marginal returns.Additionally, ICQ and FE exhibit an inherent competitive relationship, which complicates the simultaneous improvement of both. Conversely, an indiscriminate reduction in FC may impede the realization of both ICQ and FE, consequently resulting in a decline in overall TTI. Consequently, trade-offs and decisions must be made based on specific operational contexts.

For instance, for high-value, perishable agricultural products (e.g., strawberries), priority should be given to enhancing ICQ even at the expense of some FE, to ensure product quality and reliability. Conversely, for agricultural products with lower economic value and stronger storage resilience (e.g., potatoes), greater emphasis can be placed on improving FE, allowing for a certain degree of relaxation in ICQ requirements, thereby optimizing the overall system utility.

Consequently, the design of an intelligent grading system for agricultural products essentially involves solving an optimization problem: maximizing the objective function TTI under the tight constraint (FC $\leq$ $FC_{max}$).The "tight" nature of this constraint—where finite budgets cannot simultaneously achieve extremely high ICQ and FE—directly leads to the existence of an "impossible triangle" (see Section 5.1 for experimental analysis).This metric will undergo significant validation in the theoretical section of the TriAlignXA model below.

This formalized model elucidates the fundamental reason why the traditional technical approach, which pursues "absolute objective grading standards," has encountered difficulties in the agricultural products sector. It also points to the solution: a paradigm shift is imperative. The technical objective must transform from serving as the ultimate "arbitrator" to providing transparent, explainable decision-making grounds for human value judgments.This constitutes the fundamental philosophical underpinnings of the TriAlignXA framework.
\end{itemize}


\section{TriAlignXA}
\subsection{Concept Overview: A transition is underway from a paradigm of decision-makers to one of decision-support providers}

Conventional approaches to grading, predicated on the pursuit of objective standards, have endeavored to employ complex models to this end. However, these efforts are impeded by the limitations inherent to the so-called "impossible triangle. Consequently, this paper initially puts forth a paradigm shift: TriAlignXA's objective is not to function as the supreme arbiter; rather, its mission is to serve as a provider of high-quality, interpretable decision support information.In light of these findings, this paper presents the Trilemma Solver core module, which integrates three innovative engines.The overarching design objective of the framework is not to blindly pursue the maximization of a single metric (e.g., accuracy A), but rather to redefine the form of the performance function F. Through the integration of algorithmic innovations, such as feature surface modeling and process weighting, in conjunction with architectural designs, including pre-mapping mechanisms, the framework attains the global optimization of the system's overall performance P, operating within the confines of a specified stringent cost constraint Cmax.The fundamental philosophy of this approach entails a shift in the optimization objective from "absolute accuracy" to "achieving the optimal balance of interpretability, efficiency, and accuracy within limited resources."

Addressing the "impossible triangle" formed by Biological Nature(B), Perishability(T), and Cost-Effectiveness(E) is a prerequisite for designing any agricultural product grading system.Conventional single-point technical optimization has proven ineffective in addressing this challenge.

Therefore, this paper proposes that TriAlignXA is not a single model but a systematic framework.The system's fundamental component is the "Triangular Constraint Breaker," a central module that integrates three innovative engines, each designed to address a distinct aspect of the triangular constraint.

\begin{itemize}
 \item \textbf{The Biological Adaptive Engine:}
The Biological Adaptive Engine is a term used to describe a specific type of engine that has been designed to adapt to its environment in a way that is analogous to biological organisms. The concept of biodiversity (B) is incorporated, with the objective of enhancing the model's descriptive power and its capacity for generalization.

 \item \textbf{The Perishability Optimization Engine:}
The Perishability Optimization Engine is a system that has been developed to address the issue of timeliness, or the time lag between the production of agricultural products and their spoilage. The engine has been designed to ensure that the processing speed of the system remains in line with the rate of spoilage, thereby ensuring that the products are utilized in a timely manner.

 \item \textbf{The Economic Optimization Engine:}
The individual in question is responsible for the management of economic constraints (E), the objective of which is to guarantee that the entire system remains feasible within the established cost limitations.

\end{itemize}

\subsection{Three Core Engines and Six Major Strategies}

\begin{figure}[htbp]
\centering
\includegraphics[width=1.0\textwidth]{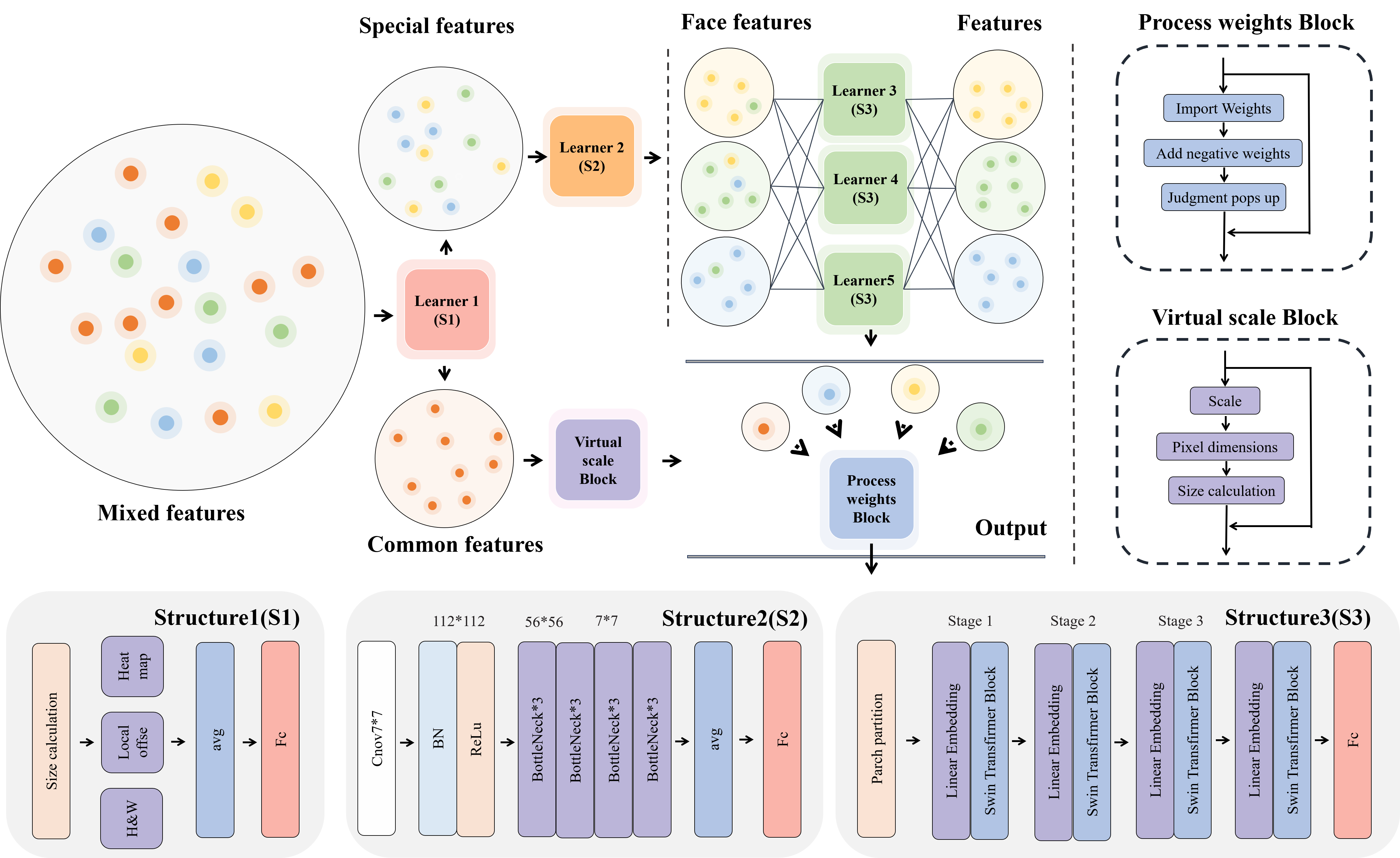}
\caption{ Partial Core Network Architecture of the TriAlignXA Framework, illustrating the dynamic transmission process of feature separation and process weighting.}
\label{fig:image3}
\end{figure}

The three core engines are designed to address the inherent biological nature, time-sensitivity, and economic characteristics of fruit and vegetable products. The TriAlignXA framework is integrated into a systematic, multi-tiered decision support system through the synergistic effects of six strategic initiatives.An illustration of part of its core network structure is shown in Figure \ref{fig:image3}.The ensuing sections will delineate the core mechanisms of each engine and the strategic implementations they encompass.

\subsubsection{\textbf{The Biological Adaptive Engine}}

\begin{itemize}
 \item \textbf{The first strategy: Regionally Graded Metric Dictionary}
Core Mechanism:
In order to address the challenges of agricultural product quality—including biodiversity (B), regional variations, and industrial economic constraints—this paper constructs a parameterized, dynamically executable intelligent central knowledge base: the Regional Graded Indicator Dictionary (RGID).The system functions not only as a repository for classification standards but also as the global parameter hub that drives the coordinated optimization of the entire TriAlignXA system.Rigid geometry (RGID) can be formally defined as a dynamic tuple whose content is uniquely determined by the origin-variety identifier ($\lambda$), as demonstrated in Equation  \eqref{eq:equ12}. 

\begin{equation}
RGID(\lambda )=\left( \Phi ,\omega ,T,W;E,D,F \right)
\label{eq:equ12}
\end{equation}

This tuple comprises two major parameter sets:

\begin{enumerate}

  \item \textbf{} Core Classification Parameter Set: It is imperative to acknowledge the role of this element in shaping fundamental classification decisions.

$\Phi$($f$;$\lambda$): Feature function set defining variety-specific feature extraction methods;
$\omega$($\lambda$): Weight vector quantifying the relative importance of each feature in composite scoring;
T($s$): Threshold function mapping continuous scores s to discrete grades;
W($\theta$;feedback): Optimization rule set dynamically updating parameters based on feedback.

  \item \textbf{} Domain-Wide Constraint Parameter Set: A Framework for Triangular Trust Index (TTI) Computation and System Optimization.

E: Economic Constraint Set, employing parameters for cost-effectiveness storage;
D: Timeliness Constraint Set, employing parameters for decay dynamics storage;
F: Trust Model Set, employing parameters for Information Coverage Quality (ICQ) computation.

The dynamic loading and execution process is outlined as follows: The system dynamically loads corresponding parameter sets based on the origin and variety labels of agricultural products.The grading process is a standardized procedure that involves extracting features based on $\Phi$, weighting scores according to $\omega$, and determining final grades.Closed-Loop Optimization: Terminal feedback, collected through Strategy Three (negative feedback loop), triggers predefined internal rules, enabling automatic adjustment of parameters like $\omega$ and continuous iteration of the knowledge base.
\end{enumerate}

Theoretical Argumentation:

For TTI, this strategy ensures consistency across the three key parameters of economics, timeliness, and trust, providing a unified data foundation for formally solving the max(TTI) optimization problem.This element serves as the cornerstone of TTI construction.

This strategy serves as the cornerstone for addressing the Biodiversity (B) challengeIt has been demonstrated to transform established agricultural prior knowledge into computable, evolvable digital standards, thereby partially alleviating the standardization dilemma exemplified by the divergent qualities of oranges in different regions, such as the south and the north, which produce fruit of varying bitter and sweet properties.By parameterizing regional variations, RGID enables a unified algorithmic framework to efficiently adapt to thousands of specialized scenarios, significantly enhancing the universality and implementation efficiency of technical solutions.When combined with negative feedback loops, this approach liberates grading standards from static constraints, allowing them to continuously evolve into a "living" system that grows alongside agricultural production practices.

 \item \textbf{The second strategy: Feature separation and feature surface modeling}
Core Mechanism:
In order to surmount the fundamental limitations of conventional models in the domain of agricultural product grading, this strategy employs a feature separation framework, as illustrated in Figure \ref{fig:image4}.

\begin{figure}[htbp]
\centering
\includegraphics[width=0.5\textwidth]{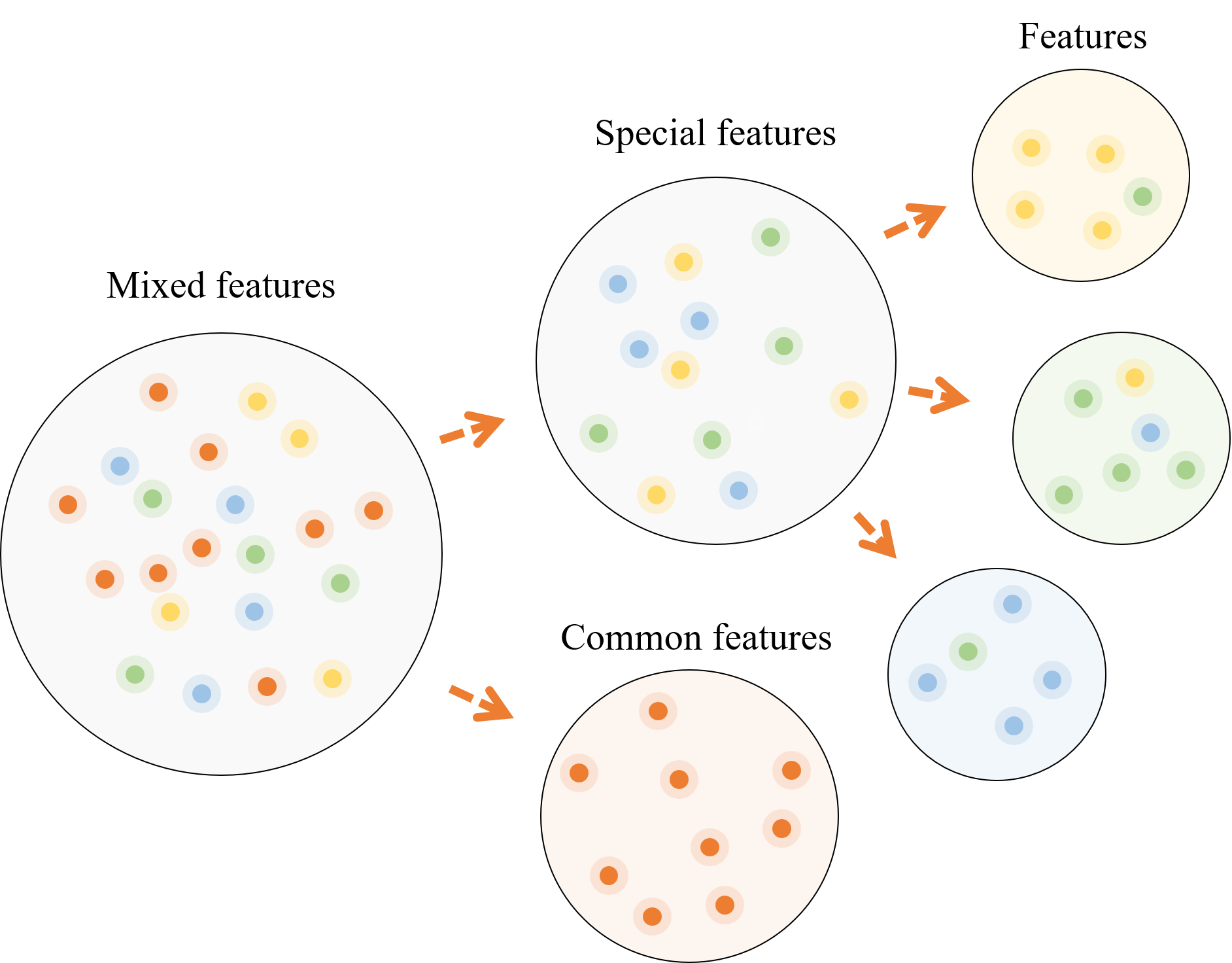}
\caption{ Feature Separation Schematic Diagram, illustrating the process of feature separation.}
\label{fig:image4}
\end{figure}

The feature separation process utilizes a dual-channel approach.The input image is represented by I. It undergoes processing through two parallel network branches, as illustrated in Equation \eqref{eq:equ13}:

\begin{equation}
\left\{
\begin{aligned}
   F_{\mathrm{general}} &= G(I;\theta_{g})  \\
   F_{\mathrm{specific}} &= S(I;\theta_{s})  \\
\end{aligned}
\right.
\label{eq:equ13}
\end{equation}

General Feature Stream: Lightweight networks have been demonstrated to extract fundamental attributes across categories, including dimension, weight, and color distribution.

A notable feature of the system under consideration is its capacity for high-capacity networking, which facilitates the capture of variety-specific characteristics, including surface texture and internal defects.

in this context, G and S represent the general feature extraction network and the special feature extraction network, respectively, with  $\theta_{g}$ and $\theta_{s}$ denoting their respective parameters.

Three-dimensional feature surface analysis:The multi-view image segmentation module decomposes the fruit into specific feature planes, simulating the observation logic of human quality inspection. The special feature stream  $F_{specific}$ is further decomposed into three feature planes: top, side, and bottom. The feature extraction process of this system is composed of three analytical modules that work in concert.The top analysis module (T) is responsible for detecting top features, such as stem integrity. The side analysis module (D) quantifies side features. The bottom analysis module (B) focuses on evaluating bottom indicators, like ripeness.As previously mentioned, the modules in question share a unified underlying parameter set $\theta$b and perform operations on the input universal feature $F_{specific}$. Ultimately, they output their respective corresponding feature surface results $F_{top}$, $F_{side}$, and $F_{bottom}$, as demonstrated in Equation \eqref{eq:equ14}:

\begin{equation}
\begin{cases}
   F_{\mathrm{top}} = T(F_{\mathrm{specific}};\theta_{s}) \\
   F_{\mathrm{side}} = D(F_{\mathrm{specific}};\theta_{s}) \\
   F_{\mathrm{bottom}} = B(F_{\mathrm{specific}};\theta_{s}) \\
\end{cases}
\label{eq:equ14}
\end{equation}

The final feature representation may be a combination of these features, as demonstrated in Equation \eqref{eq:equ15}:

\begin{equation}
{{F}_{final}}=C({{F}_{general}},{{F}_{top}},{{F}_{side}},{{F}_{bottom}},...)
\label{eq:equ15}
\end{equation}

The present study utilizes CenterNet\cite{59}as the general-purpose detector (G). The front-end network is responsible for real-time object detection and preliminary screening. It accomplishes this by employing a dynamic anchor calculation mechanism that effectively filters out background interference. Concurrently, the network extracts cross-category fundamental universal features $F_{specific}$.Subsequently, a feature analyzer (S) is introduced for the purpose of analyzing multidimensional fruit information through deep processing. This analyzer employs a dual-branch architecture, utilizing a ResNet deep network\cite{60} to deconstruct its top, side, and bottom three-dimensional feature surfaces. Concurrently, it utilizes the Swin Transformer's long-range attention capabilities\cite{61} to capture surface-specific features, such as fine textures and lesion patterns.In the final stage of the system, all multi-modal features are integrated into a lightweight multimodal fusion module (C). This module integrates feature modalities collaboratively to make comprehensive grading decisions while maintaining computational efficiency. A comprehensive exposition of the network architecture and a series of experimental comparisons can be found in the Experiments section.

Theoretical Argumentation:

A contribution to the ICQ is hereby proposed. Feature separation has been demonstrated to provide richer feature dimensions, thereby increasing the system's information coverage (SQC).Specifically, following feature separation, the system is capable of providing information across a greater number of dimensions (e.g., independent features from top, side, and bottom views), thereby increasing scq (the actual sum of provided confidence factors).Consequently, the quantity scq in the formula ICQ = min(scq / ccq, 1+$\gamma$) will increase, leading to an improvement in ICQ.

A contribution to the field of FE is demonstrated by the enhancement of computational efficiency, particularly when lightweight networks are employed for general feature streams. This enhancement is achieved through dual-channel parallel processing, which results in an increase in throughput, defined as the number of samples processed per unit time. It can be posited that, given the established relationship between throughput and the corruption rate threshold, the FE will concomitantly improve.

This strategy accomplishes the structural deconstruction of agricultural product biometric features: The model under consideration successfully addresses two significant limitations of conventional models: The terms "feature coupling" and "viewpoint confusion" are employed herein. Secondly, the practice of universal feature sharing has been demonstrated to reduce computational redundancy. Thirdly, the biometric surface and geographic dictionaries interact synergistically, thereby establishing a triangular equilibrium of "interpretability-accuracy-efficiency."

 \item \textbf{The third strategy: Negative Feedback Loop}
Core Mechanism:This strategy establishes a data-driven closed-loop optimization system, thereby enabling TriAlignXA's continuous evolution.The core of the system is predicated on the utilization of authentic end-user behavioral data (e.g., scanning QR codes, reviews, purchases, returns) as signals to iteratively refine the front-end core algorithm components.

The negative feedback loop can be formalized as a dynamic optimization equation, as demonstrated in Equation \eqref{eq:equ16}:

\begin{equation}
\theta^{(t+1)} = \theta_{(t)} - \eta \cdot \nabla_{\theta} \mathcal{L}( \mathrm{feedback} \mid \theta_{(t)} )
\label{eq:equ16}
\end{equation}

$\theta_{(t)}$: System parameters at time t;
$\eta$: Learning rate;
$L$: Loss function, quantifying the deviation between feedback and expectations.
Multi-source feedback collection: The system collects explicit user ratings and evaluations while recording behavioral data such as viewing duration for specific feature scores in "Pre-Mapping Code," repurchase rates for products across tiers, and mention frequency.The feedback dataset is defined as shown in Equation \eqref{eq:equ17}:

\begin{equation}
\mathcal{D}_{\mathrm{feedback}} = \left\{ (u_i, a_i, t_i) \mid i = 1, \dots, N \right\}
\label{eq:equ17}
\end{equation}

$u_{i}$: User behavior (scanning codes, purchases, returns, etc.), encoded as one-hot vectors;
$a_{i}$: Explicit feedback;
$t_{i}$: Implicit behavioral data (feature viewing duration, repurchase intervals, etc.).

A targeted optimization trigger has been identified. Following a thorough analysis of the feedback data, precise optimization will be initiated across various levels.

\begin{enumerate}

  \item \textbf{} Dictionary-level optimization: In instances where market feedback signifies an underestimation of a feature's importance (e.g., image shape), its weight should undergo an automatic adjustment.

  \item \textbf{} Optimization at the model level: In the event of the detection of new, unrecorded defect patterns, the initiation of targeted model fine-tuning processes is necessary to expand feature extraction capabilities.

  \item \textbf{} Rule-level optimization: In instances where the outcomes of classification exhibit systematic deviation from established market acceptance standards, prompts are generated to initiate an update to the classification thresholds contained within the threshold matrix.

\end{enumerate}

Theoretical Argumentation:

The following contributions have been made to ICQ:The negative feedback loop has been shown to enhance the system's information coverage quality (SCQ) by continuously optimizing feature weights and dynamically expanding feature dimensions. Specifically, when user behavior data indicates that the importance of a certain feature (e.g., shape) is underestimated, the system automatically increases the weight of that feature. The identification of novel defect patterns initiates the augmentation of the feature extractor.This dynamic relationship results in a continuous increase in the total sum of trust factors (scq).

The following section will examine the contributions made to FE.The negative feedback loop has been shown to enhance system processing efficiency by optimizing classification rules and model architecture.In instances where feedback data indicates systematic deviations between classification outcomes and market standards, the system simplifies decision thresholds to reduce computational complexity, thereby substantially lowering computational load.This results in an increase in the number of samples processed per unit time (throughput), which directly increases the FE value according to the formula: FE represents the throughput divided by the corruption rate threshold.

This strategy is regarded as the optimal solution for addressing the intrinsic dynamic changes in biodiversity (B).This endows the TriAlignXA framework with lifelong learning capabilities, transforming it from a static, fixed-deployment system into a living entity capable of co-evolving with agricultural production practices and market consumption preferences.This enhancement is pivotal in ensuring the system's long-term applicability and robustness.

\end{itemize}

\subsubsection{\textbf{The Perishability Optimization Engine}}

The present section is concerned with the topic of the Perishability Optimization Engine.The fundamental objective of this engine is to address the critical challenge of timeliness (T). The objective of this paper is to concentrate computational resources where they are most needed.It is imperative to ensure that the system throughput can maintain a pace that is commensurate with the rate of agricultural product spoilage.

\begin{itemize}
  \item \textbf{The fourth strategy: Cascading Grading and Process Weighting Mechanism}	
Core Mechanism:
This strategy employs a resource-aware intelligent computing architecture.At its core lies a cascaded inference pipeline integrating a dynamic decision module termed the process weighting mechanism.The recursive filtering formula is expressed in Equation \eqref{eq:equ18}:

\begin{equation}
\mathrm{Decision}(x) = 
\begin{cases}
   \mathrm{Reject}, & \text{if } S_{\mathrm{early}}(x) < \tau_{\mathrm{low}} \\
   \mathrm{Accept}, & \text{if } S_{\mathrm{early}}(x) > \tau_{\mathrm{high}} \\
   \mathrm{Continue}, & \text{otherwise}
\end{cases}
\label{eq:equ18}
\end{equation}

$S_{early}$(x): Preliminary score of sample xby the high-speed screening layer; 
$\tau_{low}$: Low rejection threshold;
$\tau_{high}$: High acceptance threshold;

Cascaded reasoning: The system does not perform calculations at uniform depth for all samples.Instead, the system utilizes a pre-screening process that involves the rapid evaluation of samples through a high-speed screening layer. This layer employs simple rules based on criteria such as brightness, size, or the weight of the product. The purpose of this preliminary screening is to swiftly eliminate products that are deemed to be severely non-compliant. Examples of products that would be considered non-compliant include items that are rotten or deformed.Subsequent samples are then subjected to increasingly sophisticated and exacting analysis.

Process Weighting Mechanism: This constitutes the intelligent core of the strategy. In the intermediate layer of the model, the process weighting mechanism must calculate the contribution value of the current feature to the final result in real time.Once the cumulative sum of contribution values exceeds the preset high acceptance threshold or falls below the low rejection threshold, the system will immediately terminate all subsequent computations for this sample and directly output the current decision result.

For samples entering subsequent analysis, the cumulative contribution sum is computed in real time at layer L, as demonstrated in Equation \eqref{eq:equ19}:

\begin{equation}
\mathrm{Decision}(x) = 
\begin{cases}
   A, & \text{if } \sum_{k=1}^{l} w_k \cdot f_k(x) \geq \tau_{\mathrm{accept}} \\
   R, & \text{if } \sum_{k=1}^{l} w_k \cdot f_k(x) \geq \tau_{\mathrm{reject}} \\
   \mathrm{Further\ Analysis}, & \text{otherwise}
\end{cases}
\label{eq:equ19}
\end{equation}

$w_{k}$: Weight of the k-th layer feature (from RGID's $\omega \lambda$)
$f_{k}$(x): Feature value of sample x at the k-th layer
A (Accept): Denotes "accept" or "qualified"
R (Reject): Denotes "reject" or "unqualified"
Further Analysis: Indicates a requirement for "further analysis" or "uncertain"

Theoretical Argumentation:

Contributions to ICQ:The system's process weighting mechanism ensures comprehensive coverage of critical quality information by analyzing boundary samples.Specifically, the recursive classification concentrates computational resources on more challenging samples with higher information value (e.g., fruits with blurred surface defects), enabling these samples to yield more comprehensive feature information (such as fruit diameter distribution and sugar core maturity). This enhancement is reflected in the SQ, which is defined as the sum of actual trust factors provided.Consequently, the scq in ICQ = min(scq/ccq, 1+$\gamma$) increases, thereby improving the overall ICQ.

The following section will examine the contributions made to FE.The hierarchical approach has been shown to enhance processing efficiency through dynamic resource allocation. Specifically, the high-speed screening layer rapidly filters out simple samples (e.g., obviously rotten or deformed fruits). The process weighting mechanism in the intermediate layer halts feature contribution calculations for clearly defined samples, executing in-depth analysis exclusively on a limited number of borderline samples.This enhancement leads to an increase in system throughput, and given the relationship between FE and Throughput, the improvement in FE is pronounced.

This strategy must be distinguished from a simplistic notion of "early exit."The system's efficacy in achieving dynamic allocation of computational resources is attributable to a process weighting mechanism, a feature that liberates valuable computing power from the handling of numerous "simple samples" and enables the concentration of efforts on the processing of those "difficult samples" situated at the boundary of classification levels.This development signifies a pivotal advancement in attaining elevated levels of accuracy and throughput within the confines of temporal constraints (T).

 \item \textbf{The fifth strategy: Separating Computation and Shelf-Life Mechanism}
Core Mechanism:
This strategy represents an innovation at the system architecture level by introducing the concept of data lifecycle management for fruits and vegetables.

Separating Computation:This approach enables the decoupling of data collection processes from computational analysis, thereby facilitating the temporal and spatial management of data.Data acquisition endpoints (e.g., origin cameras and sensors) can function in a continuous manner, with raw data being stored temporarily in buffers.The utilization of computing units enables the asynchronous extraction of data from these buffers for subsequent processing, contingent upon the availability of resources. This approach obviates the necessity for costly real-time computing hardware.

Shelf-Life Mechanism: This innovation has been meticulously designed to address the specific needs of the agricultural sector.The system assigns a biological expiration date (TTL, Time-To-Live) to each data entry.This expiration period can be dynamically configured based on the product's decay rate model (e.g., strawberry data has a short shelf life, while apple data has a longer shelf life).Subsequent to the expiration of data, the system automatically generates a flag and initiates the purging of the data.The formula for determining data validity is shown in Equation \eqref{eq:equ20}:

\begin{equation}
\mathrm{Valid}(d_i) = 
\begin{cases}
   1, & \text{if } t_{\mathrm{current}} - t_{\mathrm{collect}}(d_i) \leq \mathrm{TTL}(\lambda) \\
   0, & \text{otherwise}
\end{cases}
\label{eq:equ20}
\end{equation}

The location is as follows:$d_{i}$ is representative of a data point.The $t_{collect}$ function is used to denote the collection timestamp.TTL($\lambda$) is the biological validity period determined by the variety identifier $\lambda$.

The reduction ratio, designated as $\Delta C$,  which quantifies the decrease in storage and computational costs, exhibits a positive correlation with the proportion of expired data.The cost reduction quantification model is expressed in \eqref{eq:equ21}:

\begin{equation}
\Delta C = -\eta \cdot \frac{\sum_{i=1}^{N} \left| 1 - \mathrm{Valid}(d_i) \right|}{N}
\label{eq:equ21}
\end{equation}

In this equation, the cost coefficient, denoted by $\eta$, is expressed as the RMB per unit of sample, where as N signifies the total data volume.The primary source of system cost waste is the storage and processing of invalid data, defined as data that is no longer valid due to its expiration.The degree to which cost savings are achieved is directly proportional to the total volume of invalid data.This cost reduction is directly reflected in the decrease of FC.

Theoretical Argumentation:

Contributions to FE:Split computation is a system that enables the processing of backlogged data during periods of low load. This approach is designed to prevent computational congestion during periods of peak demand, thereby enhancing the average throughput of the system.The expiration of data cleanup (TTL mechanism) has been demonstrated to reduce redundant computations, thereby freeing computational resources for the processing of valid samples and enhancing effective throughput. The expiration of data cleanup (TTL mechanism) has been demonstrated to reduce redundant computations, thereby freeing computational resources to process valid samples and increase effective throughput.Consequently, FE is significantly enhanced.

Contributions to FC: A notable reduction in the total cost of ownership (TCO) is achieved by decoupling computation from data shelf-life management. Specifically:The concept of decoupled computation is a significant paradigm in the field of machine learning. Asynchronized data collection and processing have been demonstrated to reduce peak performance requirements for real-time computing hardware, thereby lowering initial hardware investment and energy consumption costs.Shelf-Life Mechanism: Each data point is assigned a biologically valid Time-To-Live (TTL) that is dynamically set based on a decay rate model. The automatic purging of data upon expiration ($t_{current}$ -$t_{collect}$ $>$ TTL) has been demonstrated to significantly alleviate long-term storage burdens and redundant computational loads.This reduction in the full lifecycle cost per sample has a direct impact on the FC value, leading to a substantial decrease in overall costs.

This strategy is designed to address economic constraints (E) in a clever manner.The implementation of separate computation has been demonstrated to result in a reduction of peak hardware performance requirements, thereby leading to a decrease in hardware costs.The shelf-life mechanism is designed to proactively address the time-dependent decay of agricultural data value by expiring obsolete data. This reduction in long-term storage requirements and computational demands leads to a decrease in the total cost of ownership (TCO) from a different perspective. This approach fulfills two primary objectives: "cost reduction" and "efficiency enhancement."

\end{itemize}

\subsubsection{\textbf{The Economic Optimization Engine}}

The primary objective of this engine is to regulate the hardware, computational, and storage costs (economic efficiency) of the entire system.

\begin{itemize}
  \item \textbf{The fifth strategy (Shared):}
This mechanism has been demonstrated to enhance efficiency and reduce long-term storage costs through its data expiration policy. Furthermore, the compute-disaggregated architecture eliminates the necessity for costly real-time computing hardware, consequently imposing stringent constraints on the system's total cost of ownership (TCO) from both perspectives.

  \item \textbf{The sixth strategy: Model Repository and Transfer Learning}
Core Mechanism:
To address the diverse categories and varying standards of agricultural products, this strategy designs a systematic solution for knowledge reuse and rapid adaptation.

Model Repository: The establishment of a centralized repository of pre-trained models and feature knowledge is imperative.The warehouse's fundamental layer encompasses general-purpose foundational models that have been trained on extensive datasets, such as feature extractors that have been trained on substantial fruit and vegetable datasets. The intermediate layer is responsible for storing models that have been optimized for various major categories, such as fruits and leafy vegetables. The top layer is associated with the variety parameter sets from the Regional Graded Indicator Dictionary (RGID) in Strategy One.

Theoretical Argumentation:

Contributions to FC:This strategy has been demonstrated to result in a substantial reduction in model development and adaptation costs, as evidenced by its utilization of a model repository and a transfer learning mechanism.Specifically, the model repository establishes a centralized, hierarchical knowledge base of pre-trained models, thereby eliminating the substantial overhead of training models from scratch for each new variety or region. The integration of transfer learning with lightweight fine-tuning mechanisms facilitates the adaptation to novel scenarios. This process entails the initial invocation of pre-trained models from the repository as a starting point, followed by the refinement of top-level parameters using a limited number of local samples. This approach has been shown to result in a substantial reduction in computational resources, time costs, and data annotation expenses necessary for model training, thereby significantly decreasing the per-sample development cost of models. This phenomenon directly impacts the FC value, thereby affecting the overall efficiency of the system.Transfer learning and lightweight fine-tuning: In instances where a system must adapt to a novel variety or region, there is no necessity for complete retraining from the beginning.Instead, it automatically matches or combines the most relevant pre-trained models from a model repository as a starting point, requiring only lightweight fine-tuning with a small number of local samples.This refinement typically involves updating less than 5\% of the model parameters, while achieving optimal performance.

\end{itemize}

\subsection{Integration of Pre-Mapping Mechanism and Trust Engine}

The TriAlignXA framework's core trust output mechanism, Pre-Mapping, efficiently converts technical value into commercial trust through multi-engine collaboration.This mechanism integrates feature quantification data generated by the Bio-Adaptive Engine—including 3D feature surface scores, sugar content predictions, and defect distribution heatmaps—with dynamic decision trajectories recorded by the Timeliness Efficiency Engine. These include process weighting termination points and feature reanalysis counts. In the context of economic constraints, this data is encoded into verifiable digital certificates, thereby systematically addressing the deficiencies in the quality layer of the online trust pyramid.

At the technical implementation level, the system first employs an information entropy-based feature selection algorithm for lightweight information compression, retaining the most interpretable key parameters to reduce data redundancy.Subsequently, cryptographic encapsulation generates a structured data digest containing the product's unique ID, discretized biometric encoding, dynamic decision path, and timestamp.

In the consumer engagement phase, QR codes are employed to store hash values and product identifiers. Consumers have the option to scan the codes to access lightweight, cloud-based pages, thereby retrieving multidimensional information, including three-dimensional visualizations of characteristic surfaces, product quality verification details, and safety assurance data.This innovative approach has been shown to enhance technological transparency by transforming black-box decision-making into verifiable reasoning pathways. The system replaces single-tier labeling with multidimensional feature quantification reports, thereby eliminating information asymmetry. Furthermore, it has been demonstrated that the compensatory effect for sensory experiences is achieved through sugar content/texture prediction visualization, thereby systematically reconstructing the quality trust framework.

Theoretical Argumentation:

Contributions to ICQ: The Pre-Mapping Mechanism constitutes a fundamental element of the trust infrastructure, integrating multi-source feature data generated by the TriAlignXA engine. The utilization of information entropy compression algorithms enables the encoding of high-dimensional features into lightweight data summaries, thereby facilitating the generation of verifiable, structured credentials.This mechanism has been demonstrated to significantly expand the dimensions and depth of trust factors the system can provide, thereby substantially increasing the total sum of trust factors (SCQ).According to the definition of information coverage quality: ICQ is expressed as the minimum of SQ/CQ and 1 + $\gamma$. An increase in SQ directly corresponds to higher ICQ values.Consequently, the Pre-Mapping Mechanism effectively translates the advantages of underlying algorithms into consumer-facing, perceptible trust information, systematically filling the structural gap in the quality layer of the trust pyramid.

The innovative nature of this mechanism is evident in its successful establishment of a comprehensive value loop for agricultural e-commerce through a three-dimensional transformation of "technical credibility, economic viability, and perceived experience." From the producer's perspective, it facilitates the attainment of premium brand status. From the platform's viewpoint, it mitigates the financial burden associated with quality dispute resolution. And from the industry's vantage point, it instigates a positive feedback loop wherein the possession of quality data engenders consumer trust, which in turn prompts value reinvestment.In essence, it converts the intrinsic biodiversity of agricultural products into diversified trust credentials, thereby resolving the "impossible triangle" and establishing a novel business ecosystem that prioritizes trust.


\section{Experiments and Results}

In order to provide a comprehensive validation of the theoretical model and technical framework proposed in this paper, this chapter conducts three sets of experiments in sequence: First, empirical validation of the trust pyramid model is performed; second, ablation experiments are conducted to verify the effectiveness of each core module in TriAlignXA; finally, comparative experiments are carried out to evaluate TriAlignXA.

\subsection{Validation of the Trust Pyramid Model}
The trust pyramid model posited in this study—comprising quality, safety, and marketing layers—functions as the core theoretical framework for understanding the trust dilemma in the context of fruit and vegetable e-commerce. In order to rigorously validate the hierarchical structure and its inherent "coupling" logic, this chapter abandons traditional single-dimensional verification methods and innovatively employs dual-source data comparison analysis for empirical testing. The two datasets under consideration both originate from the same large-scale consumer survey targeting the relevant population. Collectively, these studies provide substantial and readily interpretable evidence in support of the Trust Pyramid Model, with their analysis unfolding across two distinct dimensions. The terms "absolute demand" and "coupled perceived importance" are employed herein.

\subsubsection{\textbf{Research Design and Data Collection}}

Data Sources: The present study's analysis is based on two complementary sets of survey data.

The absolute demand data is measured using the multiple-choice selection rate. This data was derived from 257 multiple-choice questionnaires, wherein respondents were prompted to identify all factors that influenced their purchasing decisions from a provided list. The results are presented as the percentage of respondents selecting each option, thereby reflecting users' raw, uncompromised preferences and core concerns, uninfluenced by platform supply capability.

The second component of the data set under consideration is the Coupling Perception Data, which is represented by the mean average of a 7-point scale. This data was derived from 241 seven-point scale questionnaires, in which respondents evaluated the importance of each factor in their decision-making process (1 = not at all important, 7 = extremely important). The results are presented as average scores, reflecting the actual perceived importance of each factor under existing platform information supply conditions. This outcome is the result of the interplay between user demand and the supply capabilities of the platform.

Validation Logic: The fundamental premise of the Trust Pyramid theory is the presence of a hierarchical structure, wherein the quality (Q) of the information is prioritized as superior to the substance (S) and the medium (M). The chi-square test is employed to verify the strict presence of this hierarchy, while dual-source data will validate it from two angles:

The primary objective of this study is to ascertain whether, in the absence of external interference, users place a higher value on quality and safety. To this end, absolute demand data will be examined to determine whether users prioritize these qualities over other considerations.

Secondly, the coupling data will verify the following hypothesis: In the presence of distorted online conditions, has the focus of user perception shifted? This shift prompts the question of whether it confirms the "marketing layer substitution" dilemma.

The resulting "tension" and "paradox" from the comparison of these two elements will directly validate the "coupling gap" and "inverted pyramid" structure proposed in Chapter 3.

\subsubsection{\textbf{Hierarchical Structure Validation}}

The fundamental premise of the Trust Pyramid Model posits that consumers exhibit notable variations in the significance they attribute to the three components—Quality (Q), Safety (S), and Marketing (M)—during the decision-making process, with the components' importance decreasing in a descending order $(Q > S > M)$. To rigorously validate this hierarchical structure, this study employed a more stringent method—Cochran's Q test. This method is particularly well-suited for the analysis of response data from the same group of respondents to multiple related dichotomous variables (yes/no choices). In comparison with conventional chi-square goodness-of-fit tests, it demonstrates superior capacity to account for inherent associations among responses, thereby yielding more reliable conclusions.

The data were derived from a large-scale consumer survey involving 257 respondents. The survey employed a multiple-choice format, requiring each respondent to select all factors influencing their purchasing decisions from four provided options:The data were derived from a large-scale consumer survey involving 257 respondents. The survey employed a multiple-choice format, requiring each respondent to select all factors influencing their purchasing decisions from four provided options:

The safety assurance, designated as (A), corresponds to the safety layer, designated as S.
The quality assurance (QA) process is equivalent to the quality layer (Q).
Brand assurance, as indicated by the (C) designation, corresponds to the Marketing layer, designated as (M).
The following section is dedicated to the presentation of additional assurances.

The responses of each participant were transformed into four binary variables, with 1 representing selection and 0 representing non-selection. In conclusion, the complete response data was extracted for analysis. The null hypothesis (H0) and the alternative hypothesis (H1) for Cochran's Q test are as follows:
H0: The proportion of consumer selections across the four options is equivalent, indicating that all four options are equally significant.
H1: It is evident that a substantial discrepancy exists in the proportion of consumer selections for a minimum of two options, thereby indicating that these options possess varying degrees of importance.

Cochran's Q statistic is used to determine whether the distributions of a binary variable are consistent across multiple related samples. Its calculation is based on the total number of times each option (treatment) is selected ($G_{j}$) and the total number of options selected by each respondent ($L_{i}$). The calculation formula is shown in Equation \eqref{eq:equ22}:

\begin{equation}
Q = \frac{(k-1) \left( k \sum_{j=1}^{k} G_j^2 - \left( \sum_{j=1}^{k} G_j \right)^2 \right)}{k \sum_{i=1}^{N} L_i - \sum_{i=1}^{N} L_i^2}
\label{eq:equ22}
\end{equation}

In this context, k = 4 denotes the number of options (A, B, C, D). N=257 denotes the number of valid respondents.$G_{j}$ is defined as the total number of times the jth option was selected. $L_{i}$ denotes the total number of options selected by the ith respondent.

Results and Interpretation:

The total number of selections for each option is indicated below. The following numerical values are associated with the respective quantities: $G_{A}$=187, $G_{B}$=211, $G_{C}$=55, $G_{D}$=32. The sum of the squares of the number of options selected per respondent is 1,143, and the total number of options selected is 485. Substituting these values into the formula yields a Q-value of approximately 372.39. In the context of the null hypothesis, the Q statistic is approximately distributed as a chi-squared distribution, with 3 degrees of freedom. A consultation of the chi-squared distribution table indicates that its p-value is less than 0.001.

Cochran's Q test yielded a Q value of 372.39 with df=3 and p $<$ 0.001, indicating significant differences. Post-hoc analysis of selection frequencies confirmed a strictly decreasing relationship $(Q > S > M)$. The quality layer serves as the foundation for building consumer trust, the safety layer provides essential safeguards, while the marketing layer delivers value-added benefits. Specific results are shown in Table~\ref{tab:table1}. Dual-source data provides comprehensive and compelling evidence for this hypothesis, with partial dual-source data shown in Table~\ref{tab:table2}.

\begin{enumerate}

  \item \textbf{} The initial step in the procedure is the validation of absolute demand data. There is a broad consensus that quality and safety are fundamental principles.

Absolute demand data, as measured by multiple-choice question weighting, has been shown to indicate that quality and safety factors are of paramount importance in the decision-making process of users.The quality layer is of paramount importance. The attribute of "appearance" emerged as the paramount criterion, garnering an overwhelming 88.33\% of the total vote. In a close second, "flavor" secured 82.10\% of the votes, a notable distinction in the rankings. This unequivocally establishes that quality information is a fundamental and non-negotiable demand for users.The Safety Layer: The factor labeled "Pesticide Residues" exhibits a leadership position with a percentage of 83.66\%, a figure that exceeds most elements within the marketing layer to a substantial degree. This finding indicates that safety is regarded as an indispensable baseline requirement for users.

The Conclusion: This data offers substantial validation of the fundamental design of the Trust Pyramid, underscoring the indispensability of the Quality and Safety layers in establishing trust.

  \item \textbf{} Coupled Sensory Data Validation: Formation of an "Inverted Pyramid" Structure.

The analysis of user data, measured on a 7-point scale, reveals a distorted state of perceived importance among users under current platform conditions.

Within the quality layer, "freshness" (5.84) and "taste/flavor" (5.72) persist as the highest-scoring attributes. However, it is crucial to note that these high ratings are derived from supply imbalances within the platform. Users express a strong desire for these qualities, yet the platform is unable to fulfill their expectations.The relatively low score for "Pesticide residue testing" (4.98) in the safety layer does not indicate user indifference. Rather, it suggests diminished credibility due to insufficient quality information support.The remarkably elevated scores for "After-sales service" (5.58) and "User reviews" (5.56) in the marketing layer suggest that platform resources and user focus are being diverted toward after-sales and review mechanisms designed to compensate for underlying deficiencies.

  \item \textbf{} Dual-Source Data Comparison and Model Validation.

A comparison of the two datasets yielded the following validation conclusions:
Absolute demand data confirms the inherent structure of the pyramid (Q and S hold greater absolute importance than M) and the hierarchy $(Q > S > M)$. The integration of disparate data sets unveils the underlying online structural dynamics, with the M layer exhibiting an atypically pronounced importance, thereby reducing the disparity with the S layer. This discrepancy between the "intended" and "actual" structures not only validates the hierarchical framework but also profoundly confirms the emergence of an inverted pyramid distortion. It has been demonstrated that users do not prioritize quality and safety; rather, they are compelled to allocate greater attention to compensatory measures at the marketing layer due to the online environment's failure to effectively deliver this information.

\end{enumerate}

\subsubsection{\textbf{Comprehensive Conclusion and the Significance of Model Validation}}

Through comparative analysis of dual-source data, this study not only validates the hierarchical structure of the Trust Pyramid Model $(Q > S > M)$ but also compellingly reveals its distorted form in online environments and its underlying causes.

\begin{itemize}
 \item \textbf{}Model Validity: Absolute demand data firmly establishes that the Quality and Security layers form the absolute foundation of user trust, while the Marketing layer serves as a value-added component, validating the theoretical framework of the pyramid.

 \item \textbf{}Existence of the Coupling Gap: Paradoxes between coupling data and absolute demand data (e.g., high security demand but low perceived importance) confirm a significant coupling gap between user needs and platform offerings.

 \item \textbf{}Empirical validation of the"inverted pyramid" structure: The high scores for marketing layer factors in coupling data vividly illustrate the current state of resource misallocation and the upward shift in trust-building priorities, providing the most direct evidence for the core assertion of a "top-heavy inverted pyramid."
\end{itemize}

The significance of this validation approach lies in its transcendence of traditional hypothesis testing. It not only proves "what is" (the hierarchical structure) but also profoundly explains'why'(coupling gaps cause structural distortion) and "how" (an unstable inverted pyramid emerges). From a consumer perception perspective, this conclusion provides the most direct and intuitive theoretical basis and demand orientation for the subsequent TriAlignXA framework, which focuses on addressing the quality information black hole and the impossible triangle.

\subsection{Experimental Setup}
\subsubsection{\textbf{Multivariate Indicator Graded Dataset: Fruit3}}

\begin{figure}[htbp]
\centering
\includegraphics[width=1.0\textwidth]{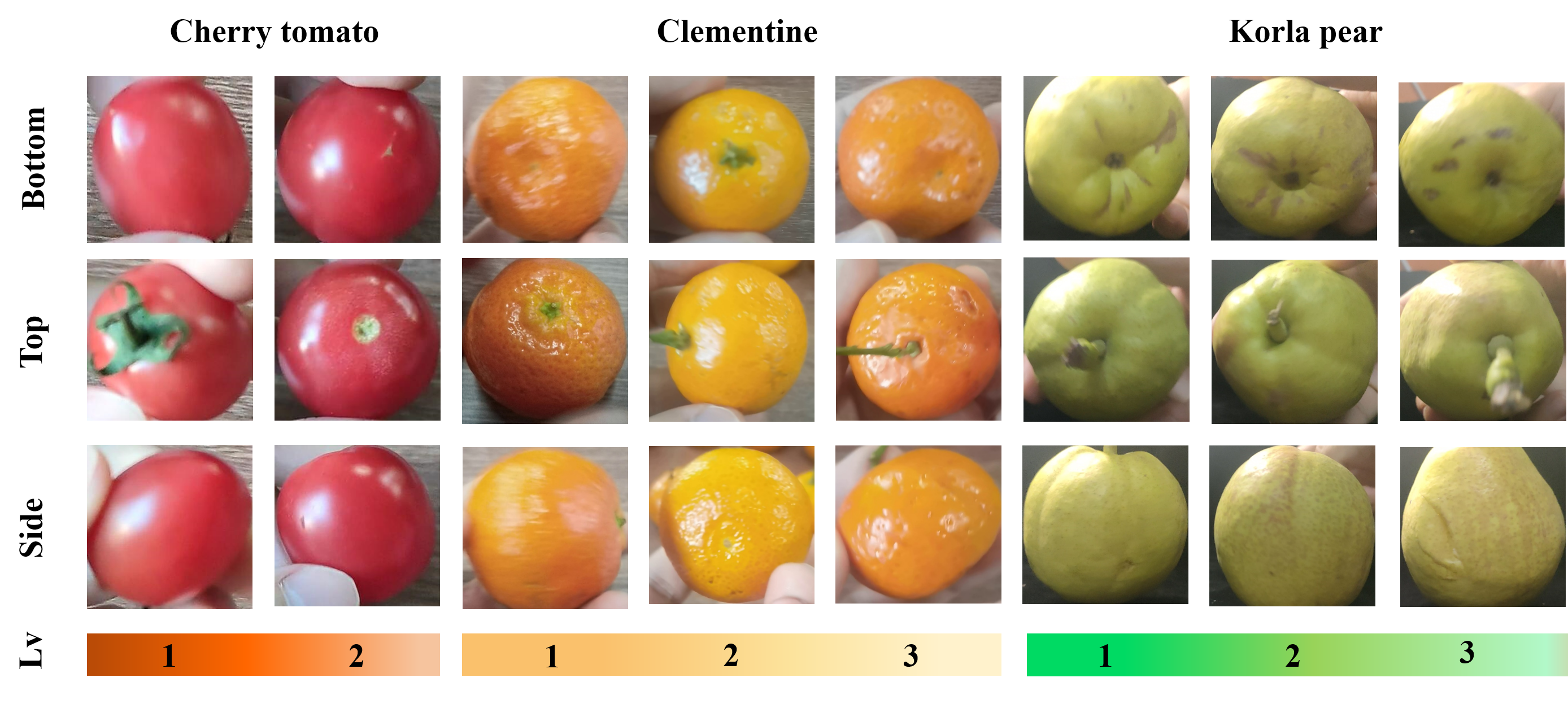}
\caption{Grading Samples in the Dataset: Cherry tomatoes, Clementines, and Korla pears. Each category is graded according to region-specific standards, specifically: cherry tomatoes follow a two-tier grading system (Premium/Grade One), while Clementines and Korla pears adopt a three-tier grading system (Grade A/B/C).}
\label{fig:image5}
\end{figure}

In order to rigorously validate the proposed TriAlignXA framework and its core strategy for addressing the fundamental constraints of the "impossible triangle", this study employs the Fruit3 dataset.As demonstrated in Figure  \ref{fig:image5}, this dataset was meticulously designed and constructed as a specialized benchmark for multimodal agricultural product grading tasks. The text explicitly addresses biodiversity, regional standard heterogeneity, and the need for rich, quantifiable feature representations. The dataset is currently available for download on Mendeley Data\cite{64}.

\begin{itemize}
 \item \textbf{}Core Design Philosophy and Principles:
In accordance with the Triangular Dilemma, Fruit3 has been meticulously engineered to symbolize the fundamental challenges inherent to the triangular dilemma. The structure of the platform is characterized by the presence of diverse fruit types, explicit geographical criteria, and multimodal characteristics. This structure functions as a realistic testing platform for the evaluation of TriAlignXA's biological adaptability, efficiency optimization, and cost-sensitivity mechanisms.

Transcending Monomodal Approaches: Fruit3 surpasses conventional pure visual classification datasets in this field by integrating authentic multimodal information crucial for holistic grading, thereby challenging models' ability to comprehensively process diverse inputs.

Reflecting Regional Standards: It is imperative to note that the grading annotations employed in this study are not arbitrary nor defined subjectively by researchers.As illustrated in Table~\ref{tab:table3} , these standards are derived in accordance with established guidelines and are publicly accessible for each fruit type.This anchors the dataset in real-world practices and mitigates the limitations of subjective grading criteria noted in prior research regarding scale.

 \item \textbf{}The present study selected three fruit varieties with significant economic value and diverse structural characteristics:
Korla Pear: The subject of representation is pome fruits that exhibit specific surface defects and shape standards.The grading process adheres to the provisions stipulated in DB 65/T 4295-2020, a document that constitutes the Xinjiang Local Standard\cite{56}.

Cherry tomatoes: These fruits, which are of the small berry variety, exhibit sensitivity to factors such as color uniformity and the presence of surface blemishes.The classification system employed in this study adheres to the provisions outlined in DB46/T 412-2016, a standard established by the Hainan Provincial Local Standard\cite{57}, which is noteworthy for its implementation of a distinctive two-tier classification framework.

Clementine: This specimen is representative of citrus fruits, with an emphasis on the integrity of the stem and the condition of the peel.The grading of these products adheres to the provisions set forth in NY/T 869-2004, a standard established by the China Agricultural Industry Standard\cite{58}.

Each fruit type comprises 150 independent samples, which were randomly collected post-harvest from representative local markets or orchards to capture natural biological variation.Per Sample:A total of twenty high-resolution images are captured from multiple preset angles (e.g., top, side, bottom) under controlled, consistent lighting conditions to comprehensively document surface characteristics.Concurrently, the system captures synchronized, high-precision weight measurement data (employing calibrated digital scales with ±0.01g accuracy), thereby establishing the core non-visual modality.The dataset under consideration contains 9,000 meticulously annotated high-resolution images (three types×150 samples/type×20 images/sample), accompanied by 150 precise weight records for each fruit type (totaling 450 weight records).

 \item \textbf{}Addressing Variability and Standardization:

The following definition of geographical standards is provided for reference: By grounding ground truth labels in specific geographical standards, the dataset explicitly encodes the inherent "geographical specificity" of agricultural classification. This addresses the challenge of "inconsistent classification standards" and provides the foundation for testing TriAlignXA's Regional Classification Indicator Dictionary (RGID).Capturing Biodiversity: A random sampling of 150 specimens per type was employed to ensure representation of natural variations in size, shape, color, and defect patterns within each fruit category.Structured Feature Representation: Multi-angle imaging strategies facilitate the decomposition of fruit into distinct feature planes, providing raw data for TriAlignXA's feature separation and feature plane modeling strategies to learn and operate upon.

 \item \textbf{}Annotation Methodology and Granularity:

The following annotation is expert-led: The execution of annotation tasks is entrusted to trained personnel who possess specialized knowledge in the domain of agricultural product quality assessment. These tasks are overseen by domain experts.Adherence to citation-based standards is paramount in ensuring consistency and objectivity.

The following annotation content is of a particularly rich nature: Each image is annotated with the following:Surface features/labels: The process entailed precise localization and identification of features, as well as the presence and integrity of fruit stems, and regions of interest.Comprehensive Grade Label: The final grade is assigned in accordance with the relevant regional standards (Korla pears and small citrus fruits: The following grades are available: A/B/C. Cherry tomatoes are also available. Extra Grade/Grade 1).This label integrates all evaluated attributes.Key Attribute Indicators: Supplementary annotations are defined as quantitative or specific elements that provide critical information as outlined in established standards. These elements may include, but are not limited to, approximate scar area classification for pears, percentage of intact pedicels for small citrus fruits, and color uniformity ratings.Quality Control: Approximately 10\% of annotations were cross-validated by a second independent annotator. To ensure the reliability of the annotations, inter-annotator reliability metrics were calculated.

The Fruit3 dataset is not merely a collection of images and labels; it is a more extensive data set.It is a purpose-built benchmark dataset designed to embody the core challenges of the "impossible triangle" in agricultural grading.The foundation for evaluating biodiversity is rooted in real-world geographical standards, integrates multimodal data, and employs structured feature representations.The following text is intended to provide a comprehensive overview of the subject matter.

 \item \textbf{}TriAlignXA's capacity to process region-specific features (via RGID) is a notable 
 \item \textbf{}The efficacy of feature separation and feature surface modeling in decomposing complex biometric features.strength.
 \item \textbf{}The following investigation will explore the efficiency gains that result from
hierarchical reasoning and process weighting under real-world constraints.
 \item \textbf{}The framework's overall accuracy and generalization capability under diverse and standardized conditions must be assessed.

\end{itemize}

\subsubsection{\textbf{Training Configuration}}

This experiment was conducted on an NVIDIA GeForce RTX 3090 (Linux system) environment.The experiment utilized the Fruit3 dataset proposed in this paper, which was divided into training, validation, and test sets at an 8:1:1 ratio.Employing a freeze training strategy to enhance resource utilization, the core parameter configuration is shown in Table~\ref{tab:table4}:

\subsubsection{\textbf{Evaluation Metrics}}

The study's pioneering contribution is the repositioning of the core paradigm of intelligent agricultural product grading from traditional "deterministic classification" to "explainable decision support." The evaluation of grading information should serve as one of the bases for decision-making, providing guidance rather than serving as the primary factor in decision-making. In light of these findings, the TriAlignXA architecture is proposed as a systematic solution to the "impossible triangle" problem, which is characterized by the confluence of "biological nature, time sensitivity, and economic viability." Furthermore, it innovatively introduces a theoretical evaluation metric—the Triangular Trust Index (TTI).At this stage, the experiment employs widely used classification performance metrics, including accuracy (ACC), precision (P), recall (R), and the F1 score (calculated as shown below). These metrics are primarily based on the following considerations:

Firstly, these metrics serve as widely adopted evaluation benchmarks in the field of intelligent classification, facilitating comparisons with existing research within a consistent framework to ensure the comparability and reproducibility of results.The present study's objective is to validate TriAlignXA's capacity to generate "quality metrics" with high discriminative power and reliability, which can effectively reflect improvements in the model's classification performance.

Second, the proposed Triangular Trust Index (TTI) serves as a system-level evaluation framework. Its theoretical model and formula derivation have been completed in the preceding sections, and its methodological validity in reconciling the "impossible triangle" has been demonstrated.However, the practical evaluation of TTI is a systematic engineering endeavor, and its parameters—such as Information Coverage Quality (ICQ), Processing Efficiency (FE), and Financial Cost Ratio (FC)—require reliance on actual hardware environments and large-scale deployments to yield reliable data.Under the current algorithm verification environment, conditions for comprehensive empirical validation are not yet available.

Furthermore, efficiency metrics such as inference latency and number of parameters measured in pure software simulation environments often fail to accurately reflect their actual performance in embedded terminals.The performance of a system is constrained by a multitude of factors, including but not limited to hardware drivers, storage bandwidth, input/output latency, and parallelization strategies. Consequently, it is more appropriate to conduct a comprehensive evaluation during subsequent system integration phases, incorporating system-level metrics such as throughput and total cost of ownership (TCO).

In summary, the selection of ACC/P/R/F1 as core evaluation metrics at this stage aims to focus on verifying the effectiveness and reliability of TriAlignXA at the algorithmic level, ensuring the assessment remains targeted and rigorous.The research is meticulously structured into two phases: theoretical validation and system implementation. The present phase has established a substantial basis for the viability of the theoretical framework and fundamental algorithms. The subsequent phase of hardware deployment and field application will entail two primary tasks: the comprehensive validation of TTI and the system-level evaluation of its economic viability and timeliness.

\begin{equation}
ACC=\frac{TP+TN}{TP+TN+FP+FN}\
\label{eq:equ23}
\end{equation}

\begin{equation}
Precision=\frac{TP}{TP+FP}\
\label{eq:equ24}
\end{equation}

\begin{equation}
Recall=\frac{TP}{TP+FN}\
\label{eq:equ25}
\end{equation}

\begin{equation}
F1=\frac{2\times Precision\times Recall}{Precision+Recall}\
\label{eq:equ26}
\end{equation}

TP: a positive sample that was correctly predicted as a positive example.
TN: a counterexample that was correctly predicted to be a counterexample.
FP: a counter-sample that was incorrectly predicted to be a positive example.
FN: positive samples that were incorrectly predicted to be counterexamples.

\subsection{Ablation Experiments}
The present study proposes a progressive ablation experiment scheme to systematically validate the effectiveness and necessity of the three core engines and six strategies within the TriAlignXA framework.The experiment was conducted on the Fruit3 dataset. The method of controlled variables was employed to introduce each core module sequentially, beginning with the baseline model, in order to quantify the contribution of each component to the dissolution of the "impossible triangle." The engine model is illustrated in Table~\ref{tab:table5} .

\subsubsection{\textbf{Baseline Model}}
The present study utilizes a ResNet-34 model that has been pre-trained on ImageNet as a robust baseline model. The model is then fine-tuned to adapt to the three-class classification task.This model is a universal solution that does not incorporate any design addressing the "impossible trinity" for agricultural products.

\subsubsection{\textbf{The Biological Adaptive Engine}}
The present study commences with an evaluation of two fundamental strategies employed by the bioadaptive engine.

\begin{itemize}
 \item \textbf{}Feature Separation: A dual-channel feature separation architecture is introduced on top of the baseline (Strategy 2).As demonstrated in Table~\ref{tab:table6} , the feature separation strategy results in an improvement of 0.85\% in ACC, with a more substantial increase of 2.76\% in Precision.This finding suggests that the separation of general and specific features effectively filters out noise, thereby enhancing the certainty of the model's judgments.

 \item \textbf{}Feature Surface Modeling: The subsequent presentation will offer a comprehensive overview of the recent advancements in three-dimensional feature surface analysis.This approach led to a substantial enhancement in recall, with an observed increase of 5.19\%. This finding underscores the efficacy of the proposed strategy in amplifying the model's capacity to discern diverse biological characteristics, particularly in the identification of stem integrity (top surface), surface defects (side surfaces), and ripeness (bottom surface).As demonstrated in Table~\ref{tab:table7} , the top layer exhibits the most significant contribution to the final decision-making process, aligning with the extant knowledge of agricultural experts.
\end{itemize}

\subsubsection{\textbf{The Perishability Optimization Engine}}
This paper builds upon the integrated bioadaptive engine, introducing the time-efficiency engine's hierarchical reasoning and process weighting mechanism (Strategy 4).This development signifies a substantial advancement in the realm of performance enhancement, with the ACC attaining a noteworthy 14.44\% increase.This strategy utilizes dynamic allocation of computational resources, facilitating expeditious decision-making for straightforward samples while concentrating computational power on complex ones. This approach has been demonstrated to substantially reduce average inference time while maintaining high accuracy, directly addressing the timeliness (T) constraint.

\subsubsection{\textbf{The Economic Optimization Engine and Pre-Mapping Mechanism}}
The TriAlignXA framework was introduced in its totality. This framework encompasses the model repository and transfer learning strategy (Strategy Six) of the economic optimization engine, as well as the pre-mapping mechanism for building trust.The complete model demonstrated an 85.87\% accuracy rate and an 87.40\% F1 score, indicating a high level of precision and effectiveness in its predictions. Although Strategy 6 did not demonstrate a direct numerical improvement, its core value lies in its ability to significantly reduce the cost of adapting to new product categories or regions while meeting the economic (E) constraint.

The experimental results in Table~\ref{tab:table6}  provide definitive evidence that TriAlignXA's three components function collectively and in a cohesive manner.The Bio-Adaptive Engine addresses the question of "how to describe," thereby elevating the upper limit of model capabilities. The Timeliness Efficiency Engine tackles the question of "how to achieve efficiency," ensuring system practicality. The Economic Optimization Engine resolves the question of "how to implement," controlling the total cost of ownership (TCO).The collaborative efforts of these three factors have systematically resolved the "impossible triangle" dilemma in agricultural product grading.

\subsection{Comparative Experiments}

Given that TriAlignXA's value lies primarily in paradigm shifts, and its weakened classification results position it more as classification-oriented information, this paper primarily conducts comparative experiments against a series of representative classical models on the Fruit3 test set to validate its continued ability to achieve favorable results in current general tasks. The selected baseline models encompass a variety of design philosophies:

Lightweight Architecture: MobileNet-v2\cite{62}, representing an efficiency-focused model. Classic deep architectures: ResNet-18, ResNet-34, and VGG-16\cite{63} are widely recognized as the prevailing standards in the field of computer vision.Cutting-edge Transformer Architectures: Swin-small is a representation of the most recent advancements in the domain of vision.All comparison models were trained end-to-end using the same training strategy as TriAlignXA (hyperparameters, data augmentation, training-validation-test split) to ensure fairness.

The experimental results are presented in Table~\ref{tab:table8} .TriAlignXA demonstrated superior performance in comparison to all baseline models across the four evaluation metrics. It attained an accuracy of 85.87\%, which represented an increase of 18.78 percentage points compared to the second-best model, Swin-small.This substantial advantage underscores the limitations of relying exclusively on deeper neural networks, more advanced general-purpose architectures, or larger parameter counts in addressing the inherent challenges of agricultural product grading within the existing paradigm.

Generalization Capability: TriAlignXA's high Recall value indicates its superior ability to handle biological diversity (B) with lower false negative rates.

Efficiency Balance: By implementing a variety of strategies, TriAlignXA substantially reduces the number of fully computed hierarchical samples, thereby markedly enhancing hierarchical efficiency.

The following section will discuss the architectural advantages of the aforementioned structure. While Swin Transformer demonstrates robust performance on general datasets, its results on Fruit3 indicate that a generic model devoid of domain-specific knowledge embedding encounters challenges in attaining optimal outcomes in this specialized task.

The comparative experiments strongly demonstrate that TriAlignXA not only achieves exceptional performance in general grading tasks, but also provides a reusable framework conceptually for solving similar challenges in the agricultural domain.

\section{Discussion and Conclusion}
\subsection{Discussion}

The prevailing focus in contemporary agricultural AI research on enhancing accuracy metrics comes at the expense of considering practical constraints.This study posits that the inherent value of technology does not lie in the pursuit of an elusive "absolute correctness" standard, but rather in the provision of comprehensive, transparent, and timely data support for human value judgments within stringent constraints.This paper proposes a novel conceptualization of agricultural product grading as an open-ended decision support task, thereby establishing a methodological framework termed "quality evidence generation–transmission–verification." This fundamental shift entails a transition from an era of "algorithms replacing humans" to a new paradigm of "algorithms augmenting humans."This approach is in alignment with the most recent developments in the field of human-machine collaborative intelligence.

The fundamental paradigm shift achieved in this study is the driving force behind the redistribution of agricultural value through the democratization of technology. In this context, algorithms transition from their traditional role as "authoritative arbitrators" to that of "evidence providers." This enables producers to gain pricing dominance based on objective quality data while empowering consumers to make autonomous decisions informed by comprehensive information. In essence, it establishes a novel equilibrium between algorithmic efficiency and human value judgments.This repositioning provides a transferable, cross-domain paradigm template for solving intelligent challenges in other highly variable biological entities, such as medical imaging diagnosis and personalized education.

The experimental results demonstrate the efficacy of TriAlignXA in mitigating the fundamental contradictions arising from the "impossible trinity" in intelligent agricultural product grading at the mechanism design level.The present framework exhibits notable advantages across three dimensions: Biological Nature, Perishability, and cost-effectiveness. The validation process has been methodically executed through the theoretical framework of the Triangular Trust Index (TTI), complemented by empirical experimentation with classical grading metrics.

In terms of Biological Nature, TriAlignXA has been demonstrated to enhance its adaptability to individual variability in agricultural products through interpretable feature separation and three-dimensional feature surface modeling.The recall rate improvement of 5.19\% is indicative of the system's exceptional performance in fine-grained quality assessment, signifying a paradigm shift from "rigid classification" to "interpretable quality descriptions." This provides a robust algorithmic foundation and reliable data support for the Information Coverage Quality (ICQ) dimension in TTI.

In terms of Perishability, the process weighting mechanism enables resource-adaptive dynamic inference scheduling, significantly reducing processing latency while maintaining high precision.This design effectively addresses the time constraints imposed by the spoilage rate of fresh produce at the system architecture level, thereby laying the technical foundation for subsequent system-level verification of the processing efficiency (FE) dimension in TTI, including throughput and response time.

In terms of cost-effectiveness, the model repository and decoupled computing architecture have been demonstrated to have a significant impact on model reuse and system operating costs.By calibrating parameters, a high degree of adaptation to diverse crop varieties and geographical regions is achieved, thereby enhancing the economic viability of the technology in practical agricultural settings. This fulfills the fundamental requirements of the economic cost ratio (FC) dimension within the TTI framework.

The present study has now fully completed the theoretical development, validation of the algorithm, and architectural design, forming a systematic research phase.Through rigorous validation of core metrics such as accuracy, recall, and F1 score, TriAlignXA has exhibited superior performance in classification tasks.The subsequent phase will concentrate on advancing hardware adaptation, large-scale deployment, and commercial environment testing to further refine the system-level validation of TTI metrics.

Subsequent research will adhere to a three-phase approach to advance industrialization and comprehensive TTI validation:
\begin{itemize}
 \item \textbf{}The present study will address the pilot deployment and causal validation. Collaborate with e-commerce platforms to conduct A/B testing, quantifying the impact of the Pre-Mapping mechanism on key business metrics such as purchase conversion rates and return rates, and refining the TTI parameter system.

 \item \textbf{}Cross-Domain Generalization and Privacy Protection: The objective of this study is to achieve collaborative optimization of regional graded indicator dictionaries (RGID) through federated learning. This approach is designed to enhance system generalization capabilities and data security.

 \item \textbf{}The deployment of the ultra-lightweight edge is a process of paramount importance. The integration of neural architecture search and model compression technologies has been demonstrated to further reduce deployment costs and expand the system's real-time processing capabilities at the edge.
\end{itemize}

In summary, TriAlignXA's innovative value lies not only in its systematic resolution of the "impossible triangle," but also in establishing a multidimensional evaluation paradigm through the Triangular Trust Index (TTI). This provides theoretical underpinnings, technical solutions, and practical pathways for building trust within the agricultural e-commerce sector.The platform's open architecture design has been shown to lower barriers to industry collaboration, thereby driving the formation and development of a new paradigm for the agricultural trust economy.

\subsection{Conclusion}

The trust crisis in online fresh food e-commerce is rooted in the asymmetry and low credibility of quality information.The fundamental limitation of traditional technological approaches lies in their failure to systematically address the constraints imposed by the "impossible triangle" formed by the interplay of biological factors (B), time constraints (T), and economic considerations (E).To this end, this study proposes a new systemic evaluation paradigm centered on the Triangular Trust Index (TTI), driving a fundamental shift in the role of technology from "arbitrator pursuing absolute standards" to "enabler providing transparent decision-making foundations." This transition signifies a fundamental transformation from an era of "algorithms replacing humans" to a paradigm of "algorithms augmenting humans."The main contributions of this study include the following four aspects:

\begin{itemize}
 \item \textbf{}The Theoretical Framework Construction and Validation: A TTI evaluation system centered on information coverage quality (ICQ), processing efficiency (FE), and economic cost ratio (FC) was proposed. Empirical research validated the foundational role of verifiable quality information in establishing online trust.

 \item \textbf{}Significant technological advancement: The TriAlignXA system framework was proposed, leveraging three core engines and six strategic approaches to systematically address the challenges of the "impossible triangle."The experimental results demonstrate that the system achieves an accuracy rate of 85.87\% in hierarchical tasks and exhibits strong potential for synergistic optimization across all dimensions of TTI.

 \item \textbf{}Innovation in Trust Transmission Mechanisms: The "Pre-Mapping Mechanism" was proposed, encoding multidimensional quality data generated during algorithmic processes into lightweight QR codes. This approach facilitates low-cost accessibility and verifiability of quality information, thereby providing a practical implementation pathway for the ICQ dimension within TTI.

 \item \textbf{}Data and Benchmark Contribution: The Fruit3 dataset was constructed and subsequently released under an open-source license, ensuring compliance with established regionalization standards. This dataset serves as a public benchmark for future research endeavors concerning intelligent grading of agricultural products and TTI-related studies.
\end{itemize}

This study demonstrates that in agricultural product quality evaluation, the core value of technology lies not in pursuing objectivity in grading standards, but rather in providing efficient, transparent, and cost-effective information support for decision-making by all relevant parties within the constraints of the "impossible triangle."The significance of TriAlignXA lies not only in its accuracy but also in its ability to optimize trust-building efficiency across multiple dimensions through TTI.In the coming years, research will continue to evolve in the following three directions:

\begin{itemize}
 \item \textbf{}It is imperative to further reduce the system's economic cost (FC), explore ultra-lightweight edge deployment, and expand practical application scenarios.

 \item \textbf{}The enhancement of information coverage quality (ICQ) can be achieved through the integration of novel sensing technologies, which facilitate non-destructive quantification of internal quality attributes, including but not limited to sugar content and acidity.

 \item \textbf{}Conduct large-scale industrial experiments to quantify the causal impact of TriAlignXA and TTI mechanisms on core business metrics (such as conversion rates and return rates) through A/B testing.
\end{itemize}

This paper proposes a systematic solution that is guided by TTI theory and powered by TriAlignXA technology. This solution will propel fresh food e-commerce toward a new trust-driven model, shifting away from cost control.The system's open, modular architecture design allows for the flexible replacement or addition of modules by developers, thereby providing a robust foundation and ongoing momentum for the digital transformation of the agricultural value chain.

\clearpage 

\appendix
\section{} 

\begin{table}[h]
\centering
\caption{Frequency of Consumer Selection for Fruit Purchase Assurance Attributes and Cochran's Q Test Results (N=257)}
\label{tab:table1}
\begin{tabular}{l r r r}
\toprule
\textbf{Attribute} & \textbf{Selection Count} & \textbf{Percentage (\%)} & \textbf{Response Percentage (\%)} \\
\midrule
Quality Assurance (Q) & 211 & 82.1 & 43.5 \\
  \addlinespace[0.2cm]
Safety Assurance (S)  & 187 & 72.8 & 38.6 \\
  \addlinespace[0.2cm]
Market Assurance (M)   & 55  & 21.4 & 11.3 \\
  \addlinespace[0.2cm]
Other Assurance       & 32  & 12.5 & 6.6 \\
  \addlinespace[0.2cm]
Total                 & 485 & 188.7 & 100 \\
\bottomrule
\multicolumn{4}{p{1.0\textwidth}}{\footnotesize \textit{Note:} Response Percentagerefers to the percentage of respondents who selected each attribute in a multiple-response question. Percentageindicates the proportion of the total sample (N=257) that selected each attribute. The total exceeds 100\% due to multiple responses.}
\end{tabular}
\end{table}

\begin{table}[ht]
  \centering
  \caption{The following table shows the network models in each network structure learner, the trained dataset features, and each of the features trained here comes from the above dataset.}
  \label{tab:table5}
  \begin{tabular}{p{1.2cm}p{2.5cm}p{2.8cm}p{7cm}} 
    \toprule
    \textbf{Learners} & \textbf{BaseNetwork} & \textbf{Features} & \textbf{Significance} \\
    \midrule
    1 & CenterNet & Species & The redundant magazines in the scene are removed by target detection to distinguish between generic and specific features. \\
    \addlinespace[0.2cm]
    2 & ResNet-18 & Posture faces & Perform attitude surface interrogation on filtered data to distinguish attitude surfaces. \\
    \addlinespace[0.2cm]
    3 & \multirow{3}{*}{Swin} & surface defects & Fruit defects are detected and then graded. \\
    \addlinespace[0.2cm]
    4 &  & Appearance & The fruit surface is tested according to standards and then graded. \\
    \addlinespace[0.2cm]
    5 &  & Stem (of fruit) & The fruit stalks are tested according to standards and then graded. \\
    \bottomrule
  \end{tabular}
\end{table}

\newcolumntype{Y}{>{\RaggedRight\arraybackslash}X}
\begin{table}[htbp]
\centering
\caption{Comparative Analysis of Selected Dual-Source Data}
\label{tab:table2}
\renewcommand\arraystretch{1.2}

\resizebox{0.88\textwidth}{!}{%
\begin{tabularx}{\textwidth}{Y Y Y Y Y}
\toprule
\textbf{Trust Layer} & 
\textbf{Key Indicator} & 
\textbf{Absolute Demand Data(\%)} & 
\textbf{Coupling Perception Data} & 
\textbf{Data Analysis and Interpretation} \\
\midrule

 & Quality Assurance & 81.71\% & 5.66 & 
\textbf{Core Conflict Point:} Extremely high user absolute demand, but \\

Quality Layer& Appearance & 88.33\% & 5.48 &ineffective online transmission leads to perception scores below  \\

& Taste/Flavor & 51.75\% & 5.72 &  expectations, revealing a significant coupling gap. \\

\midrule
 & Safety Assurance& 70.04\% & 5.16 & 
\textbf{Baseline Demand:} Users highly concern about safety, but   \\

Safety Layer& Pesticide Residue & 83.66\% & 4.98 &  the lack of visual quality information undermines the credibility of safety claims\\

& Origin Tracing & 28.02\% & 5.12 &  (e.g., text reports), resulting in relatively low perception scores. \\

\midrule
 & Marketing Assurance & 22.18\% & 5.53 & 
\textbf{Distorted High Score:} Users claim to not prioritize brand marketing the most, yet online   \\

Marketing Layer& User Reviews & -- & 5.56 &platforms concentrate resources here (after-sales, reviews). This leads to these “band-aid” services compens- \\

& After-Sales Service & -- & 5.58 &  ating for underlying deficiencies receiving abnormally high perceived importance. \\

\bottomrule
 \multicolumn{5}{p{1.0\textwidth}}{\footnotesize \textit{Note:} Absolute demand data comes from Questionnaire 1 multiple-choice question: "What aspects do you focus on more when selecting fruits usually?"; Coupling perception data comes from Questionnaire 2 scale questions. Indicators differ slightly between the two questionnaires to better align with user needs; a hyphen (-) indicates data not collected for that indicator in the other questionnaire.}
\end{tabularx}%
} %
\end{table}
\newcolumntype{P}[1]{>{\RaggedRight\arraybackslash}p{#1}}
\newcolumntype{C}[1]{>{\Centering\arraybackslash}p{#1}}

\newlength{\wFruit}   \setlength{\wFruit}{0.14\textwidth}
\newlength{\wMetrics} \setlength{\wMetrics}{0.14\textwidth}
\newlength{\wA}       \setlength{\wA}{0.24\textwidth}
\newlength{\wB}       \setlength{\wB}{0.24\textwidth}
\newlength{\wC}       \setlength{\wC}{0.24\textwidth}

\setlength{\tabcolsep}{1pt}
\renewcommand\arraystretch{1.18}
\begin{table}[htbp]
\centering
\caption{The table below shows the breakdown of grading indicators in the local policy documents for the three fruits, with three grades for Kuril pears and clementine and only two grades for Cherry tomatoes.}
\label{tab:table3}
\renewcommand\arraystretch{1.15}
\begin{tabularx}{\textwidth}{P{\wFruit} P{\wMetrics} P{\wA} C{\wB} P{\wC}}
\hline 
\multirow{2}{*}{\textbf{Fruit}} &
\multirow{2}{*}{\textbf{Metrics}} &
\multicolumn{3}{c}{\textbf{Fruit Grade}} \\
    \cmidrule{3-5}
& & \textbf{Grade A} & \multicolumn{1}{c}{\textbf{Grade B}}  & \textbf{Grade C} \\
\hline 

& Appearance
& \multicolumn{2}{P{\dimexpr \wB+\wC+\tabcolsep\relax}}{%
  The fruit shape is straight, the fruit surface is smooth and clean, fruit peduncle is complete
} & The fruit shape is upright, the fruit surface is smooth and clean, fruit stalk is intact\\

Kurla pear& Color and luster
& \multicolumn{3}{P{\dimexpr \wA+\wB+\wC+2\tabcolsep\relax}}{%
  Yellow-green, green with a reddish tinge, Crispy, juicy and refreshing
} \\

& Fruit surface defects 
  & no scarring 
  & Scarring allowed, total area of single fruit $\leq 0.8\ \mathrm{cm}^2$ 
  & Scarring allowed, total area of single fruit $\leq 1.0\ \mathrm{cm}^2$ \\

& Weight (g) 
  & $\geq 120,\ \leq 160$ 
  & $100 \leq \text{weight} < 120$ 
  & $80 \leq \text{weight} < 100$ \\
\hline 

& Appearance 
  & oblate, fruit apex slightly protruding, fruit bottom flat, consistent shape 
  & oblate, apex slightly protruding, bottom flat, relatively uniform in shape 
  & oblate, fruit apex protruding, fruit bottom flat, no obvious deformity \\

Clementine& Stem (of fruit) 
  & incomplete 
  & 95\% complete 
  & 90\% complete \\

& Diameter (mm) 
  & 45--50 & 40--45 / 50--55 & 35--40 / 55--60 \\
\hline 

& Appearance 
  & High degree, no general defects 
  & Higher degree, a few defects allowed 
  &  \\

Cherry tomato& Stem (of fruit) 
  & Fresh green 
  & slightly bright green 
  &  \\

& Weight (g) 
  & 10.6--20.6 & 13.0--20.0 &  \\

& Background (texture) 
  & Firm, elastic fruit 
  & Fruit slightly firm, slightly elastic 
  &  \\
\hline 
\end{tabularx}
\end{table}

\begin{table}[ht]
  \centering
  \caption{Experiment Configuration}
  \label{tab:table4}
  \begin{tabular}{lccccccr}
    \toprule
    I\_lr & Min\_lr & Optimizer\_type & Momentum & Weight\_decay & Lr\_de\_type & Workers \\
    \midrule
    1e-2 & L\_lr*0.01 & sgd & 0.9 & 5e-4 & cos & 4 \\
    \bottomrule
    \multicolumn{7}{p{1.0\textwidth}}{\footnotesize \textit{Note:} I-lr: Initial learning rate; Min-lr: Minimum learning rate; Optimizer-type: Type of the optimizer (SGD); Momentum: Momentum parameter; Weight-decay: Weight decay coefficient;Lr-de-type: Learning rate decay type (cosine annealing); Workers: Number of data loading threads.}
  \end{tabular}
\end{table}

\begin{table}[ht]
  \centering
  \caption{Comparative Trials}
  \label{tab:table6}
  \begin{tabular}{p{1.66cm}p{1.66cm}p{1.66cm}p{1.66cm}p{1.66cm}p{1.66cm}p{1.66cm}p{1.66cm}}
    \toprule
    \textbf{Base-line} & \textbf{B-Engine} & \textbf{P-Engine} & \textbf{E-Engine} & \textbf{ACC(\%)} & \textbf{P(\%)} & \textbf{R(\%)} & \textbf{F1(\%)} \\
    \midrule
    $\checkmark$ & -- & -- & -- & 66.24 & 72.16 & 66.24 & 69.07 \\
    \addlinespace[0.2cm]
    $\checkmark$ & $\checkmark$ & -- & -- & 67.09 & 74.92 & 67.09 & 70.74 \\
    \addlinespace[0.2cm]
    $\checkmark$ & $\checkmark$ & $\checkmark$ & -- & 71.43 & 73.34 & 71.43 & 72.24 \\
    \addlinespace[0.2cm]
    $\checkmark$ & $\checkmark$ & $\checkmark$ & $\checkmark$ & 85.87 & 88.98 & 85.87 & 87.40 \\
    \addlinespace[0.2cm]
    -- & -- & -- & -- & $\uparrow$19.63 & $\uparrow$16.82 & $\uparrow$19.63 & $\uparrow$18.33 \\
    \bottomrule
  \end{tabular}
\end{table}

\begin{table}[ht]
  \centering
  \caption{Three Eigenfaces validate the effect.}
  \label{tab:table7}
  \begin{tabular}{p{2.65cm}p{2.65cm}p{2.65cm}p{2.65cm}p{2.65cm}}
    \toprule
    \textbf{Metric} & \textbf{Top(\%)} & \textbf{Side(\%)} & \textbf{Bottom(\%)} & \textbf{Mix(\%)} \\
    \midrule
    W-P             & 77.64         & 68.32         & 67.37           & 73.34        \\
        \addlinespace[0.2cm]
    W-R             & 70.72         & 69.23         & 75.49           & 71.43        \\
        \addlinespace[0.2cm]
    ACC             & 70.72         & 69.23         & 75.49           & 71.43        \\
        \addlinespace[0.2cm]
    F1              & 74.02         & 68.77         & 71.20           & 72.24        \\
        \addlinespace[0.2cm]
    Weight          & 56.08         & 22.17         & 21.75           & --           \\
    \bottomrule
  \end{tabular}
\end{table}

\begin{table}[ht]
  \centering
  \caption{Comparative Trials of Different Models}
  \label{tab:table8}
  \begin{tabular}{p{2.59cm}p{2.59cm}p{2.59cm}p{2.59cm}p{2.59cm}}
    \toprule
    \multirow{2}{*}{\textbf{Model}} & \multicolumn{4}{c}{\textbf{Metric}} \\
    \cmidrule{2-5}
    & \textbf{ACC(\%)} & \textbf{P(\%)} & \textbf{R(\%)} & \textbf{F1(\%)} \\
    \midrule
    Mobilenet\_v2 & 63.91 & 61.70 & 63.91 & 62.79 \\
            \addlinespace[0.2cm]
    Resnet\_18 & 61.36 & 57.26 & 61.36 & 59.24 \\
            \addlinespace[0.2cm]
    Resnet\_34 & 62.21 & 59.49 & 62.21 & 60.82 \\
            \addlinespace[0.2cm]
    Swin\_small & 67.09 & 74.91 & 67.09 & 70.79 \\
            \addlinespace[0.2cm]
    Vgg\_16 & 65.18 & 68.49 & 65.18 & 66.80 \\
            \addlinespace[0.2cm]
    \textbf{Ours} & \textbf{85.87} & \textbf{88.98} & \textbf{85.87} & \textbf{87.40} \\
    \bottomrule
  \end{tabular}
\end{table}

\clearpage 
\begin{refcontext}[sorting = none]
\printbibliography
\end{refcontext}
\end{document}